\PassOptionsToPackage{numbers,sort&compress}{natbib}

\documentclass{article}

\usepackage[final]{neurips_2026}

\usepackage[final]{neurips_2026}

\makeatletter
\providecommand{\@trackname}{Preprint} 

\makeatother

\usepackage[utf8]{inputenc}
\usepackage[T1]{fontenc}
\usepackage{hyperref}
\usepackage{url}
\usepackage{booktabs}
\usepackage{amsfonts}
\usepackage{amsmath}
\usepackage{nicefrac}
\usepackage{microtype}
\usepackage{xcolor}
\usepackage{graphicx}
\usepackage{multirow}
\usepackage{subcaption}
\usepackage{enumitem}
\usepackage{pifont}
\usepackage{wrapfig}
\usepackage{caption}
\usepackage[table]{xcolor}
\usepackage{array}

\usepackage{booktabs}

\usepackage{longtable}

\usepackage{listings}

\usepackage{float}

\title{SteelBench: Evaluating Vision-Language Models\\in Real-World Industrial Environments}

\author{%
  \textbf{Suryanarayana Reddy Yarrabothula}$^{1,2}$ \thanks{Corresponding author. Email: \texttt{yarrabothula@iitbhilai.ac.in}}, \\
  \textbf{Manisha Chawla}$^1$, 
  \textbf{Gagan Raj Gupta}$^1$, \\
  \textbf{Kunal Sinha}$^3$, 
  \textbf{Sashank Lekkala}$^1$, 
  \textbf{Ashirvadhan Dosapati}$^1$, \\
  \textbf{Saikamal Nannuri}$^1$, 
  \textbf{Katragadda Ajay RamaSwamy Chowdary Gowtham}$^1$ \\
  $^1$Indian Institute of Technology Bhilai,
  $^2$Steel Authority of India Limited,
  $^3$VIT Vellore\\
  \texttt{yarrabothula@iitbhilai.ac.in}
}
\begin{document}

\maketitle
\begin{abstract}
Existing video benchmarks evaluate action recognition on consumer videos,
egocentric recordings, or simulated industrial environments. They do not test
vision-language models under the visual and procedural conditions of real
industrial CCTV, where workers appear as distant figures amid dust, steam, low
light, glare, occlusion, and overlapping activities. We introduce
\textsc{SteelBench}, a diagnostic benchmark for industrial surveillance that
jointly evaluates per-worker activity recognition, safety-rule reasoning, and
annotation provenance. SteelBench contains 1,345 densely annotated clips,
curated from 149 hours of operational plant footage and 10,024 candidate
clips using temporal deduplication, class balancing, and
visibility-aware stratified sampling. Each clip includes dense per-worker
action labels, PPE attributes, spatial context, and safety-rule annotations.

Because model-assisted annotation can shape the labels later used for model
evaluation, SteelBench includes a provenance-aware audit protocol. The protocol
measures label influence, evaluates sensitivity to ground-truth provenance, and
reports a human reference from expert-reviewed labels. Applying this audit, we
find that unaudited VLM-sourced ground truth can inflate same-family model
accuracy by up to 17 percentage points. Across nine VLMs from four architectural
families, the best model reaches only 42.6\% action accuracy, compared with an
84.6\% human benchmark. Performance also fragments across recognition,
robustness, calibration, and safety reasoning. Even when models predict the
correct action, 37-58\% of cases still yield incorrect safety judgments,
and no model passes more than 2 of 5 diagnostic checks. SteelBench shows that
real industrial activity understanding requires provenance-aware and
failure-mode-specific evaluation rather than leaderboard accuracy alone. The dataset is publicly available on \href{https://huggingface.co/datasets/steelbench}{Hugging Face}.

\end{abstract}
\section{Introduction}
\label{sec:introduction}

Vision-language models (VLMs) are routinely evaluated on action recognition,
visual question answering, and scene understanding, but predominantly on
data where subjects are clearly visible, centrally framed, and recorded
under controlled or simulated conditions. Video benchmarks such as
Kinetics~\citep{kinetics}, ActivityNet~\citep{activitynet}, and
AVA~\citep{ava} capture human activity at close range and high resolution,
while egocentric datasets such as Ego4D~\citep{ego4d} and
Assembly101~\citep{assembly101} focus on detailed hand-object interactions.
Many existing industrial video benchmarks use simulated or controlled
environments~\citep{industryeqa}, while related safety datasets often
evaluate narrower tasks, such as PPE detection from curated
images~\citep{sh17dataset, constructionsafetyai, vlmsafetyinterpretable}. None of these datasets, to our knowledge, combines dense per-worker action annotation, PPE assessment, spatial context, and safety-rule compliance from real operational industrial CCTV. Workers appear as $80$--$200$ pixel figures
at $7$--$10$ meters. Visibility conditions arise from natural plant
operations, including occlusion from worker activities and equipment
placement, low light, dust from material handling, steam from cooling, and
glare from molten metal. Multiple workers often perform different activities
simultaneously. These conditions also make annotation difficult. When visual evidence is ambiguous, annotators may defer to VLM-generated
pre-fills, creating a dependency between model-assisted label construction
and model evaluation that most benchmarks do not
audit~\citep{wang2024humanllm, datacuration2024}.
\begin{figure}[htpb]
    \centering
    \includegraphics[width=\textwidth]{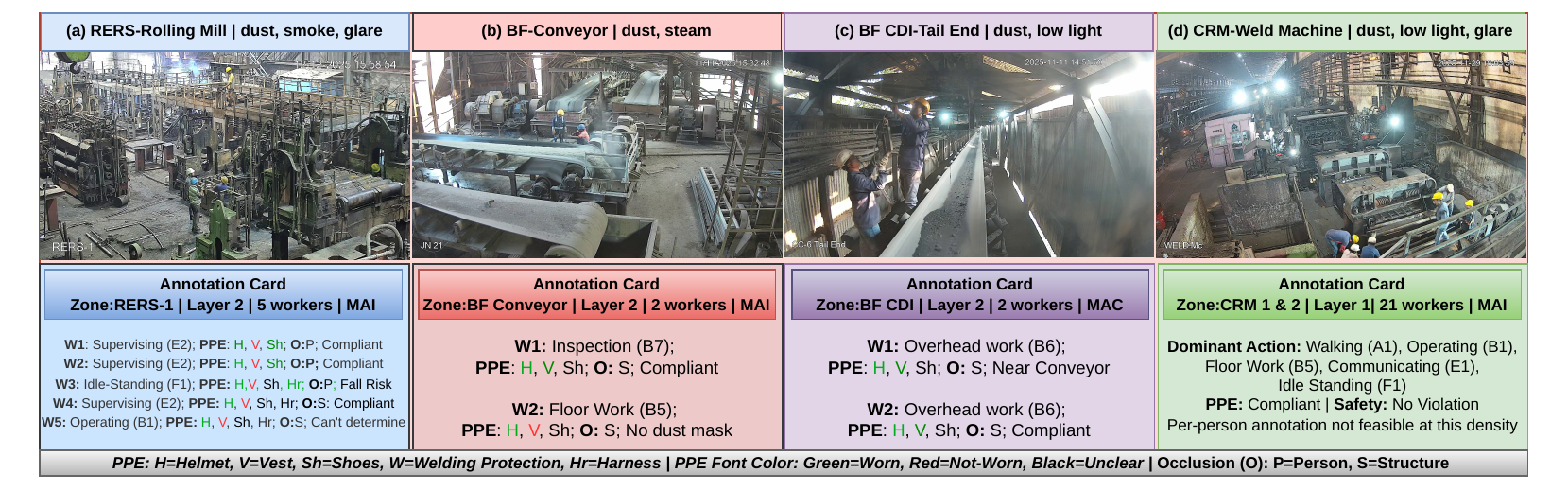}
    \caption{An illustration of SteelBench data representing real industrial surveillance conditions, with workers appearing as distant figures amid dust, steam, low light, and glare, while multiple simultaneous activities unfold across the scene, each with corresponding annotation labels. SteelBench incorporates a 2-Layer Schema ($\leq 5$ persons for Layer-2 \& $> 5$ persons for Layer-1) for annotations with per-worker labels across 25 action classes, PPE assessment, visibility tags, spatial context, and safety compliance, up to 58 labels per clip.} 
    \label{fig:overview}
\end{figure}

We introduce \textbf{SteelBench}, a diagnostic benchmark for industrial
surveillance, constructed from a single operational integrated steel plant, that jointly evaluates activity recognition, safety-rule
reasoning, and annotation provenance. In real industrial surveillance, these dimensions cannot be evaluated independently. Unreliable labels distort model comparison, and activity recognition is only meaningful when it supports reliable downstream safety judgment. To ensure
label reliability despite domain ambiguity, we introduce a provenance-aware
audit protocol (Section~\ref{sec:audit}) that measures how VLM pre-fills influence labels and downstream evaluation.

Evaluating Nine VLMs against audited ground-truth (GT) reveals that recognition, robustness, calibration, and safety reasoning fragment across model families, with no model jointly meeting the diagnostic requirements for reliable autonomous deployment(Section~\ref{sec:results}). These patterns remain stable
under taxonomy-preserving prompt and frame-sampling (Section~\ref{sec:ablations}), indicating structural capability gaps rather than prompt artefacts.

Our contributions are:
\begin{itemize}[leftmargin=1.2em]
    \item \textbf{\underline{SteelBench}}: A real industrial surveillance benchmark with dense per-worker annotations covering actions, PPE, visibility, spatial context, and safety compliance (Section~\ref{sec:dataset}).
    
    \item \textbf{\underline{Provenance-Aware Audit Protocol}}: A framework for auditing model-assisted benchmark construction and evaluation.
    
    \item \textbf{\underline{A Diagnostic VLM Evaluation}}: An evaluation across 9 models revealing fragmented action recognition, robustness, calibration, and safety reasoning capabilities.
\end{itemize}

\section{The SteelBench Dataset}
\label{sec:dataset}

SteelBench is drawn from continuous CCTV footage recorded at an operational
integrated steel plant over five months (Table~\ref{tab:dataset_summary}). Although SteelBench draws from a single facility, an integrated steel plant spans functionally distinct sub-domains with different camera geometries, lighting regimes, and activity profiles. We group 16 zones into 6 industrial domains (rolling mills, raw material handling, ironmaking, steel making, utilities, and chemical processing). Each domain contributes 16-25 action classes and distinct visibility conditions. Cross-domain accuracy varies by 8-16pp within each model. Full per-domain results appear in Appendix~\ref{app:domain_diversity}. The 1,345 clips result from a 27:1 curation of 10,024 candidate clips. Temporal deduplication removes near-identical surveillance views, and stratified sampling ensures coverage across action classes, sites, and visibility conditions.

\begin{wraptable}{r}{0.48\textwidth}
\centering
\caption{SteelBench dataset summary.}
\label{tab:dataset_summary}
\small
\begin{tabular}{ll}
\toprule
Raw footage & 149 hours \\
Source videos & 117 \\
Cameras / zones & 64 / 16 \\
Candidate clips & 10,024 \\
Final evaluation clips & 1,345 \\
Person-level instances & 2,208 \\
Layer 2 (per-person) & 805 clips \\
Layer 1 (scene-level) & 540 clips \\
Action taxonomy & 25 classes, 6 groups \\
PPE items per worker & 5 \\
Scene types & SA, MAI, MAC \\
Visibility conditions & 6 (multi-label) \\
Annotation fields per clip & 9-58 \\
\bottomrule
\end{tabular}
\end{wraptable}

Clips contain 9-58 (mean 24.7) structured annotation fields such as action code, 5 PPE items-helmet, vest, safety shoes, welding protection,
harness, spatial context, safety violations, occlusion, and transitions per worker, plus scene-level metadata and provenance tracking, yielding over 42,000 individual annotation decisions across the dataset. By comparison, Kinetics-400~\cite{Kay2017TheKH} provides a single action label per clip, whereas UCF101~\cite{Soomro2012UCF101AD} provides a single class tag per video. The total human annotation effort exceeds 160 person-hours across 2,384 submitted annotations (including 289 double-annotation pairs for IAA and 508 expert reviews), with a median per-clip annotation time of 3.9 minutes. This annotation density, combined with the provenance audit that tracks every field modification, supports the fine-grained diagnostic analyses (per-class, per-condition, per-site). Additional annotation-scale comparisons are provided in Table~\ref{app:scale_comparison}.

Each clip is represented by 8 frames at 1080p resolution, preserved because
PPE items occupy only 25-40 pixels at surveillance distance. Five trained Tier-1 annotators, one of whom had their annotations removed later to preserve the quality of data(~\ref{app:annotator10_case}), independently corrected VLM(Qwen3-VL-235B) pre-fills, and two domain experts subsequently verified each annotation without seeing the original VLM output. For each visible worker, annotators record an action code from a 25-class taxonomy in six groups: \textbf{A} Locomotion (5), \textbf{B} Stationary Work (9),
\textbf{C} Crane and Equipment (4), \textbf{D} Material Handling (3),
\textbf{E} Social (2), and \textbf{F} Idle (2). The taxonomy reflects two
constraints: (i) safety rules are defined at this granularity, and (ii) overhead CCTV
at 7 to 10 meters cannot resolve finer categories. Scenes are
categorised as SA (single actor), MAI (multi-actor independent), or MAC (multi-actor coordinated). Annotations pass through four stages: VLM pre-fill (using Qwen3-VL-235B), Tier-1 annotator correction, expert verification, and safety officer adjudication. Provenance is tracked at every stage (Section~\ref{sec:audit}).

\begin{figure}[htbp]
    \centering
    \includegraphics[width=\linewidth]{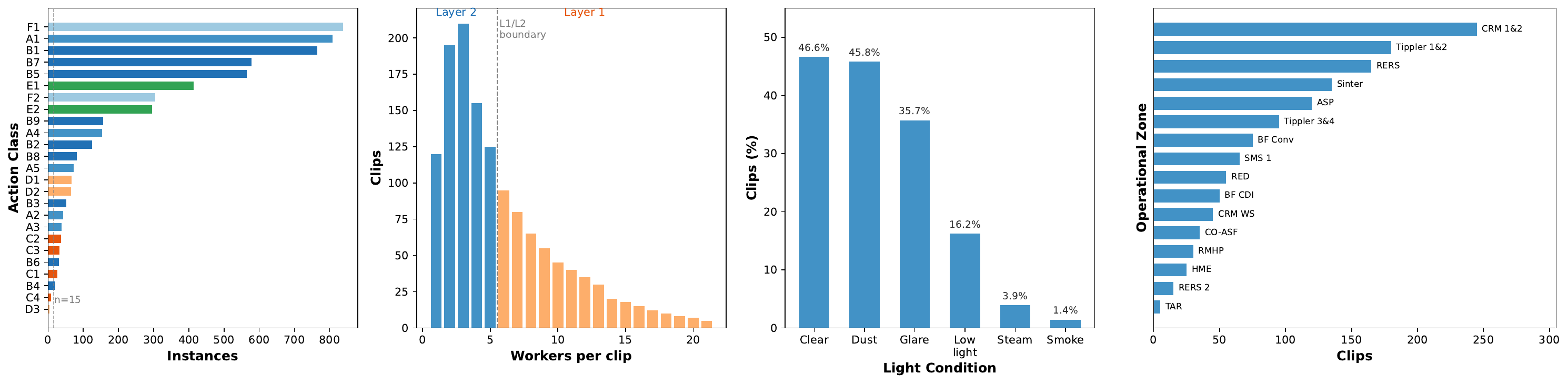}
    \caption{SteelBench data distributions. (a) Action class distribution.
    The long tail reflects genuine operations. Dashed line marks $n$=15
    threshold. (b) Worker count per clip with Layer 1/2 boundary.
    (c) Visibility conditions (multi-label). (d) Clips across 16 zones.}
    \label{fig:distributions}
\end{figure}

We retain the natural action distribution rather than artificially balancing
it (Figure~\ref{fig:distributions}). Classes with fewer than 15 instances
are excluded from per-class claims. Group-level analysis is used for
Groups C and D, where individual class support is limited. Further details on
the curation pipeline are in Appendix~\ref{app:pipeline}. Furthermore, inter-annotator agreement reaches $\kappa$=0.780 on action classification
and $\kappa$=0.793 on PPE assessment. These numbers alone do not verify
whether annotators converged independently or through shared VLM exposure.
Section~\ref{sec:audit} separates these two sources of agreement.
\section{Provenance-Aware Audit Protocol}
\label{sec:audit}

\begin{figure}[htbp]
    \centering
    \includegraphics[width=0.92\linewidth]{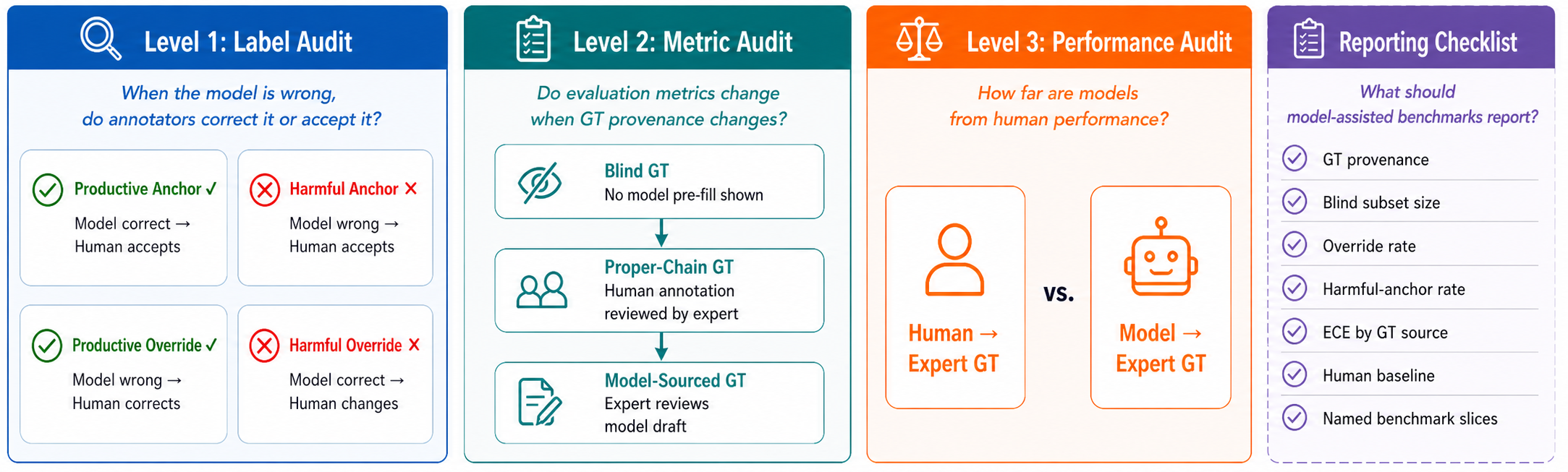}
    \caption{Provenance-aware audit protocol for model-assisted benchmark
    construction. The three levels measure label influence, metric sensitivity
    to GT provenance, and human reference performance.}
    \label{fig:audit_protocol}
\end{figure}

Model-assisted annotation introduces dependencies between benchmark construction and evaluation, particularly when models are later evaluated on labels influenced by related systems. When a VLM generates pre-fill suggestions for annotators to review, final labels may carry model-induced patterns rather than independent human judgment. If models from the same family are later scored on such labels, apparent performance can be inflated. Recent work supports \cite{preferenceleakage2025, schroeder2025humanloop, datacuration2024} this concern. Datasheets for
Datasets~\citep{gebru2021datasheets} improve dataset reporting by trying to standardize this, but do not
define provenance-specific measurements. To address this gap, we introduce a provenance-aware three-level audit protocol, summarised in Figure~\ref{fig:audit_protocol}, and apply it to SteelBench in Section~\ref{sec:audit_findings}.


\textbf{Level~1: Label influence.} Each annotator and VLM interaction is
classified against an expert reference into four outcomes. Productive anchors
accept correct VLM suggestions. Productive overrides correct VLM errors.
Harmful anchors accept VLM errors. Harmful overrides change correct VLM
suggestions. Let $A$, $O$, $H_A$, and $H_O$ denote these four counts. We define
\[
\mathrm{CLR}=\frac{A+O}{A+O+H_A+H_O},
\qquad
\mathrm{CR}=\frac{O}{H_A} \quad (H_A>0).
\]
CLR is the expert-correct interaction rate. CR measures corrected VLM errors per accepted VLM error.

\textbf{Level~2: GT provenance sensitivity}. The same model is evaluated
against three GT slices. \emph{Blind} labels are created without
VLM suggestions. \emph{Proper-chain} labels pass through VLM pre-fill, human
correction, and expert verification. \emph{VLM-sourced} labels are produced
when experts review VLM output directly. The VLM predictions are identical across all three evaluations, and only the reference labels change. Shifts in measured accuracy therefore isolate the effect of annotation provenance on evaluation outcomes, regardless of whether the underlying clips overlap across slices. 

\textbf{Level~3: Human reference.} Proper-chain human labels are compared
against the same expert reference used in Level~1. This estimates a human
reference point and inter-annotator agreement for the task.

Two additional metrics complete the protocol. Anchoring bias measures how much VLM exposure shifts annotator behaviour:
\[
\Delta_{\mathrm{anchor}}
=
P(\hat{y}_{\mathrm{human}}=\hat{y}_{\mathrm{VLM}}\mid \mathrm{anchored})
-
P(\hat{y}_{\mathrm{human}}=\hat{y}_{\mathrm{VLM}}\mid \mathrm{blind}).
\]
The override rate is the fraction of clips where the annotator changed at least
one pre-fill field. For calibration, we compute ECE separately on each GT
provenance slice using the confidence scores and binning scheme defined in
Section~\ref{sec:framework}. At minimum, model-assisted benchmarks should
report the pre-fill source, blind-sample construction, override rates, and
provenance-stratified accuracy and calibration.

\section{Evaluation Methodology}
\label{sec:framework}

\textbf{Evaluation Protocol:} We evaluate 9 VLMs spanning frontier proprietary models (GPT-4o, GPT-5.4, Claude Opus 4.7, Gemini 2.5 Pro and Flash), large open-weight models (Qwen3.5-122B, Llama 4 Maverick), and small-to-medium open-weight models (Gemma 4-31B, Nemotron-12B). Each clip comprises 8 frames at the original 1080p resolution. All models receive the same zero-shot
structured prompt containing the action taxonomy, site context, PPE definitions, and safety-rule instructions. Each model returns a structured prediction with per-worker action labels, PPE status, safety assessment, and a self-reported confidence score. 

We parse outputs using rule-based extraction with an over 98\% success rate across all models. Failed parses are treated as incorrect predictions. Full prompt text, parsing rules, and model configurations are in Appendix~\ref{app:metrics}. Predictions are compared against the GT slices
defined in Section~\ref{sec:audit}. The human reference point is established through the audit protocol (Section~\ref{sec:audit} and Section~\ref{sec:audit_findings}).

\textbf{Diagnostic Metrics:} Action accuracy and macro-F1 measure recognition performance, but miss three failure modes specific to industrial surveillance. A model may perform well under clear conditions but fail under natural industrial visibility degradation. A model may recognise an action correctly yet
apply the wrong safety rule. A model may report high confidence even when wrong, leaving human operators without a basis for triage. We address each gap using a targeted metric (Table~\ref{tab:metrics_purpose}) along with additional derived metrics described in Appendix~\ref{app:metrics}.

\begin{table}[t]
\centering
\caption{Evaluation metrics. Each addresses a failure mode that standard
accuracy does not capture.}
\label{tab:metrics_purpose}
\small
\begin{tabular}{ll}
\toprule
\textbf{Evaluation question} & \textbf{Metric} \\
\midrule
Can the model recognise worker activity? & Accuracy, Macro-F1 \\
Does performance hold under plant conditions? & nAUDC \\
Does correct perception produce correct reasoning? & CRG \\
Is confidence meaningful for human review? & ECE \\
Does the model satisfy joint requirements? & DRS \\
\bottomrule
\end{tabular}
\end{table}

\textbf{nAUDC} (Normalised Area Under the Degradation Curve) summarises
accuracy across six visibility conditions, normalised by the model's
best-condition accuracy:
\[
\mathrm{nAUDC} = \frac{\frac{1}{|V|}\sum_{v \in V} \mathrm{Acc}_v}
{\max_{v \in V}\; \mathrm{Acc}_v}
\]
We normalise by the best condition rather than the mean because the mean
would mask condition-specific failures. nAUDC of 1.0 means no
condition-dependent drop.

\textbf{CRG} (Compositional Reasoning Gap) isolates safety reasoning failures from perception failures:                                                                                                                                                                        
  \begin{equation}                                                                                                                                                                                          
  \mathrm{CRG} =                                                                                                                                                                                            
  1 - \frac{\sum_i \mathbb{1}[\hat{a}_i = a_i \;\land\; \hat{s}_i = s_i]}                                                                                                                                   
           {\sum_i \mathbb{1}[\hat{a}_i = a_i]},                                                 \label{eq:crg}                                                   
  \end{equation}                                                                                                                

where $\hat{a}_i, a_i$ are predicted and GT action codes and $\hat{s}_i, s_i$ are predicted and GT safety judgments. The denominator restricts evaluation to instances where the model correctly identifies the action, isolating reasoning failures from perception failures.

\textbf{ECE} (Expected Calibration Error) measures whether predicted confidence reflects actual accuracy. Because VLM confidence is self-reported rather than derived from log-probabilities, we treat ECE as a reliability
diagnostic for human-in-the-loop triage.

\textbf{DRS} (Deployment Readiness Score) aggregates five diagnostic checks into a single indicator: action separability (DWA $\geq$ 0.80), robustness (nAUDC $\geq$ 0.85), reasoning (CRG $\leq$ 0.20), violation detection (safety recall $\geq$ 0.90), and worker detection (F2-detect $\geq$ 0.70). DRS is not a regulatory certification but is derived from plant safety officers' inputs based on their minimum deployment requirements and should be treated as a diagnostic checklist showing which capabilities are present and which are missing. Threshold sensitivity is analysed in Section~\ref{sec:ablations}. Component definitions are in Appendix~\ref{app:metrics}.

\section{Experimental Results \& Discussion}
\label{sec:results}

We evaluate 9 VLMs spanning frontier proprietary and open-weight families.
We first apply the audit protocol from Section~\ref{sec:audit} to
characterise label reliability and provenance effects, then report model
evaluation findings.

\subsection{Audit Findings}
\label{sec:audit_findings}

Table~\ref{tab:audit_level1} and Table~\ref{tab:audit_level23} summarise the audit results from applying the
three-level protocol defined in Section~\ref{sec:audit} to SteelBench. Table~\ref{tab:audit_level1} shows that VLM \textbf{Qwen3-VL-235B} pre-fills substantially influence annotator behaviour during label construction, but the effect is predominantly corrective rather than imitative. Most interactions either preserve correct VLM suggestions or involve annotators overriding incorrect ones, while harmful anchoring remains comparatively rare. This pattern suggests that the three-tier verification pipeline successfully limits error propagation from model-assisted pre-fills and preserves high agreement with expert references despite heavy VLM involvement.

\begin{table}[htpb]
\centering
\begin{minipage}[t]{0.48\textwidth}
\centering
\caption{Level 1 audit metrics for SteelBench direction analysis. Direction analysis uses 2,829 field-level comparisons from 177 proper-chain clips
against an expert reference.}
\label{tab:audit_level1}
\small
\begin{tabular}{ll}
\toprule
\textbf{Audit metric} & \textbf{Value} \\
\midrule
Productive anchoring & $65.8$\% \\
Productive override & $27.0$\% \\
Harmful anchoring & $3.9$\% \\
Harmful override & $3.3$\% \\
Correct-label rate (CLR) & $92.8$\% \\
Correction ratio (CR) & $6.9$$\times$ \\
\bottomrule
\end{tabular}
\end{minipage}
\hfill
\begin{minipage}[t]{0.48\textwidth}
\centering
\caption{Level 2 and Level 3 audit metrics for provenance sensitivity
and human reference evaluation. Per-field and per-annotator breakdowns are in
Appendix~\ref{app:audit_details}.}
\label{tab:audit_level23}
\small
\begin{tabular}{lc}
\toprule
\textbf{Audit metric} & \textbf{Value} \\
\midrule
Blind GT accuracy & $37.2$\% ($n$=102) \\
Proper-chain GT accuracy & $57.4$\% ($n$=169) \\
VLM-sourced GT accuracy & $77.7$\% ($n$=157) \\
\midrule
Human accuracy & $84.6$\% \\
Expert IAA ($\kappa$) & $0.82$ \\
\bottomrule
\end{tabular}
\end{minipage}
\end{table}

The provenance gradient shows why audit matters even when harmful anchoring
is low. Evaluating \textbf{Qwen3.5-122B} against annotations assisted by its same-family model, \textbf{Qwen3-VL-235B}, inflates apparent accuracy on the reference label sets constructed under three different conditions: blind ($n$=102), proper-chain ($n$=169), and VLM-sourced ($n$=157). The VLM reports roughly 90\% confidence in all three cases, yet measured accuracy shifts from 37.2\% to 57.4\% to 77.7\%. We treat full-GT accuracy as a ranking signal and use blind and proper-chain slices for provenance-sensitive claims. 

\textbf{Human reference:} We measure human reference accuracy by comparing Tier-1 action labels against expert-verified labels on 370 person-level pairs from 174 proper-chain clips. Tier-1 annotators agree with expert judgment on 84.6\% of action labels ($\kappa$=0.82). This establishes the task's human reference for surveillance distance and camera resolution. Full methodology and per-annotator breakdowns appear in Appendix~\ref{app:human_baseline}.

The audit protocol also detected low-quality annotation in practice. One annotator (annotator\_10) exhibited an override rate of 16\%, VLM agreement of $\kappa$=0.775 (vs.\ team range 0.43-0.59), and zero productive overrides in direction analysis. These signals indicated rubber-stamping of VLM pre-fills rather than independent review. The annotator's 41 clips were excluded from the final ground truth, thereby improving all aggregate quality metrics. (Appendix~\ref{app:annotator10_case}).

\subsection{Model Evaluation}
\label{sec:model_eval}
Table~\ref{tab:multimodel} presents the main results. The key findings are discussed one by one.

\begin{table}[htpb]
\centering
\caption{Main results across 9 VLMs on SteelBench. Action accuracy is
computed over 1,345 clips. CRG is computed on the action-correct person-level
instances. Checks indicate the number of diagnostic thresholds passed out
of five (DWA, nAUDC, CRG, safety recall, F2-detect). Human reference
accuracy is 84.6\% ($\kappa$=0.82).}
\label{tab:multimodel}
\small
\setlength{\tabcolsep}{4pt}
\renewcommand{\arraystretch}{1.15}
\begin{tabular}{l l  c c c c c c}
\toprule
\textbf{Tier} & \textbf{Model} & \textbf{Acc$\uparrow$} & \textbf{F1$\uparrow$} & \textbf{nAUDC$\uparrow$} & \textbf{CRG$\downarrow$} & \textbf{ECE$\downarrow$} & \textbf{Checks} \\
\midrule
\rowcolor{blue!6}
Frontier & GPT-4o           & 38.8 & 22.3 & 0.858 & 0.390 & 0.469 & 2/5 \\
\rowcolor{blue!6}
         & GPT-5.4          & 35.5 & 16.1 & 0.684 & 0.476 & 0.399 & 1/5 \\
\rowcolor{blue!6}
         & Claude Opus 4.7  & 37.4 & 17.8 & 0.806 & 0.573 & \textbf{0.286} & 1/5 \\
\rowcolor{blue!6}
         & Gemini 2.5 Pro   & 29.5 & 17.8 & 0.788 & 0.495 & 0.590 & 1/5 \\
\rowcolor{blue!6}
         & Gemini 2.5 Flash & 33.6 & 17.3 & 0.848 & 0.496 & 0.562 & 1/5 \\
\midrule
\rowcolor{green!6}
Large open & Qwen3.5-122B     & \textbf{42.6} & \textbf{22.6} & 0.806 & 0.582 & 0.475 & 1/5 \\
\rowcolor{green!6}
           & Llama 4 Maverick & 34.5 & 12.7 & 0.909 & \textbf{0.375} & 0.502 & 2/5 \\
\midrule
\rowcolor{orange!8}
Small/med & Gemma 4-31B      & 39.2 & 16.2 & \textbf{0.912} & 0.551 & 0.453 & 2/5 \\
\rowcolor{orange!8}
          & Nemotron-12B     & 27.8 & 17.0 & 0.813 & 0.520 & 0.605 & 1/5 \\
\bottomrule
\end{tabular}
\end{table}

\textbf{Finding 1: VLMs remain far below human performance.} The best model
(Qwen3.5-122B) reaches 42.6\% action accuracy against the 84.6\% human
reference. This 42pp deficit holds across all 9 models. Macro-F1 is
substantially lower at 12.7 to 22.6\%, confirming that common classes
inflate accuracy while rare classes are missed. On the related simulated
industrial benchmark IndustryEQA, the same frontier models score higher. For
example, GPT-4o reports 57.2\% on IndustryEQA~\citep{industryeqa}, compared
with 38.8\% on SteelBench. Although the tasks differ, the gap supports the
need for evaluation on real operational footage.

\begin{figure}[htbp]
    \centering
    \includegraphics[width=\linewidth]{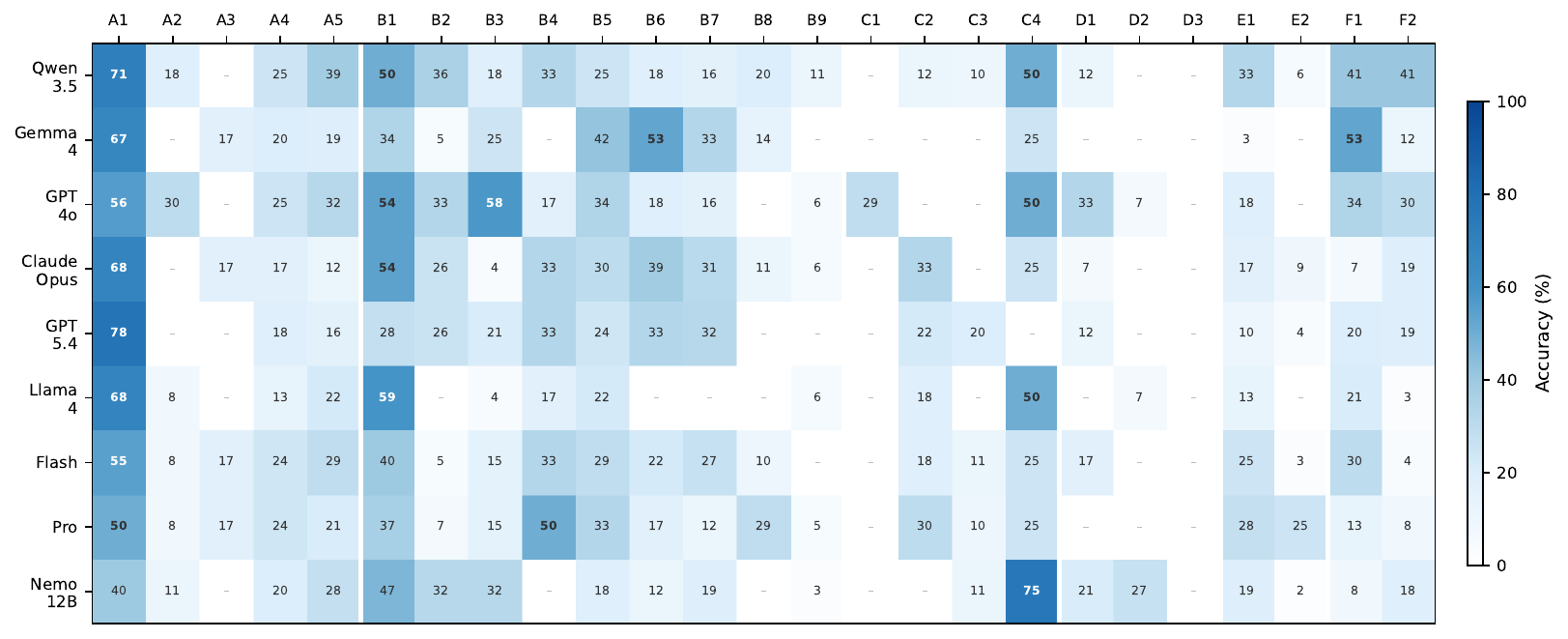}
    \caption{Per-class accuracy (\%) across 25 action classes and 9 models.
    Action groups separated by white dividers: \textbf{A}~Locomotion (A1--A5),
    \textbf{B}~Stationary Work (B1--B9), \textbf{C}~Crane \& Equipment (C1--C4),
    \textbf{D}~Material Handling (D1--D3), \textbf{E}~Social (E1--E2), \textbf{F}~Idle (F1--F2).
    Groups C and D remain difficult despite adequate support. Classes with $n<15$ (C4, D3) are shown but excluded from per-class
  claims. B1 is brighter
    because multiple models default to it when uncertain.}
    \label{fig:perclass}
\end{figure}

\begin{wrapfigure}{r}{0.55\textwidth}
    \centering
    \vspace{-10pt}
    \includegraphics[width=0.55\textwidth]{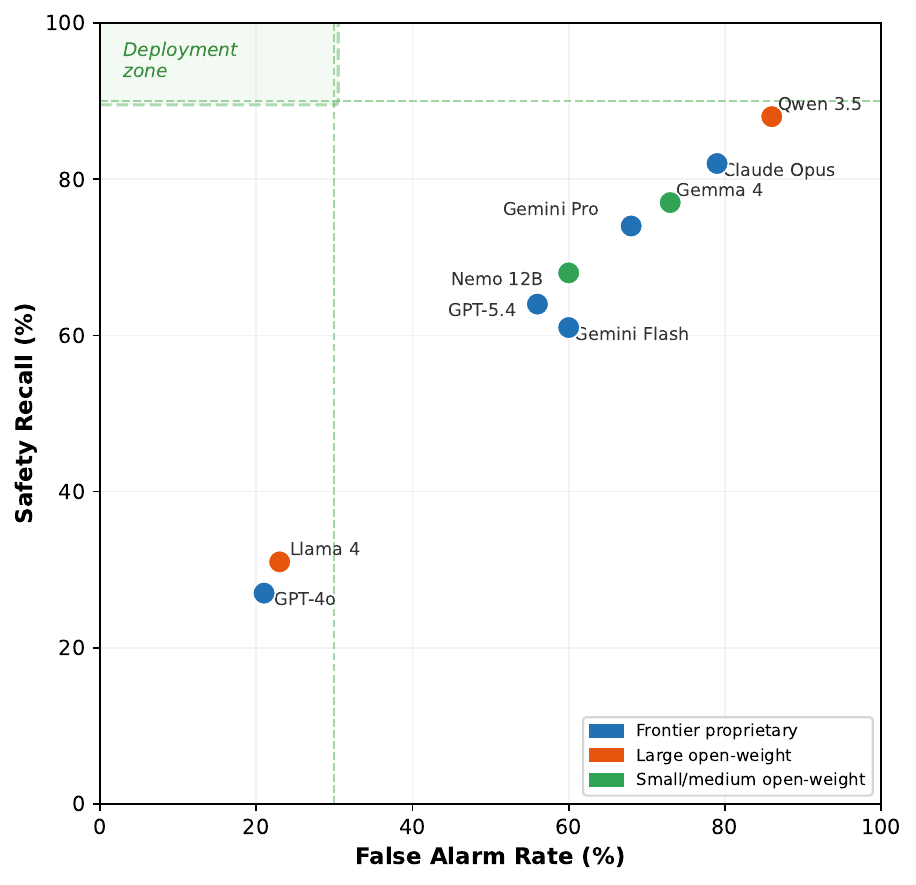}
    \caption{Safety recall versus false alarm rate across 9 models. The shaded region requires at least 90\% recall and at most 30\% false alarm rate. No
    model falls inside this region.}
    \label{fig:safety_scatter}
    \vspace{-10pt}
\end{wrapfigure}

\textbf{Finding 2: Failure is structured by action group and model family}
The per-class heatmap (Figure~\ref{fig:perclass}) shows that Groups A and B
are partially recognisable, with an accuracy range of 49-79 pp, while Groups C (crane operations) and D (material handling) have an accuracy of 10-75\% 7-33\% at the group level (130 and 147 instances, respectively) due to small sample size for some classes. Per-class support within these groups is limited, so we report group-level results as noted in Section~\ref{sec:dataset}. These low scores are structured rather than random. Five of nine models assign B1 (operating machinery) to 16 to 28\% of their errors. Gemma defaults to F1 (idle), while Claude and GPT-5.4 default to B7 (inspection). At surveillance distance, a worker near the equipment is difficult to distinguish from one operating it, and each model family resolves this ambiguity differently. Detailed per-class counts and per-site breakdowns are in Appendix~\ref{app:full_results} and~\ref{app:site_results}.

\textbf{Statistical reliability:} Of 25 action classes, 23 have $n \geq 15$ instances. The two classes below this threshold (see Figure~\ref{fig:distributions}) are excluded from per-class analysis.
 Bootstrap confidence intervals ($\pm$2.1pp at 95\%, 10,000 resamples) confirm that the 42.0pp human-model gap exceeds sampling noise by a factor of 10 and that model rankings are stable (Appendix~\ref{app:bootstrap}). DRS pass/fail outcomes are robust to $\pm$10\% threshold variation (Section~\ref{sec:ablations}).        

\textbf{Finding 3: Accuracy does not predict robustness or calibration.}
Gemma 4 achieves lower action accuracy than Qwen3.5-122B (39.2\% vs.\ 42.6\%) but substantially better robustness (nAUDC 0.912 vs.\ 0.806), while GPT-5.4 shows the sharpest degradation (0.684). Most models remain heavily overconfident, reporting 75--90\% confidence at only 28--43\% actual accuracy; Claude Opus 4.7 is the main exception (ECE 0.286). Across all diagnostic checks, no model demonstrates consistently reliable behaviour, and performance does not systematically correlate with model scale or access tier.


\textbf{Finding 4: Correct action recognition does not guarantee correct
safety reasoning.} CRG ranges from 0.375 (Llama~4) to 0.582
(Qwen3.5-122B), computed over 597 to 930 action-correct person-level
instances per model. This means 37 to 58\% of correctly labelled actions
still produce wrong safety judgments.
The highest-accuracy model also exhibits the worst reasoning gap (CRG 0.582), indicating that correct action perception does not guarantee correct downstream safety judgment. Safety failures further separate into two family-level regimes: false-alarm-biased and false-safe-biased models (Figure~\ref{fig:safety_scatter}).
In seven models, 60 to 93\% of safety
errors are false alarms. In GPT-4o and Llama~4, 63 to 66\% of safety errors
are missed violations. No model achieves both a high recall and a low false
alarm rate. This split is consistent across action groups and visibility
conditions (Appendix~\ref{app:false_safe}).
Section~\ref{sec:ablations} tests whether these findings are sensitive to
prompt and frame-sampling choices.


The sim-to-real gap further underscores the need for real-footage benchmarks. GPT-4o achieves 57.2\% on the simulated IndustryEQA benchmark~\citep{industryeqa} but only 38.8\% on SteelBench, an 18.4pp drop. This gap confirms that simulated-industrial evaluation overestimates deployed capability (Appendix~\ref{app:sim_real}).
\section{Ablation Study}
\label{sec:ablations}

\begin{figure}[htbp]
    \centering
    \includegraphics[width=\linewidth]{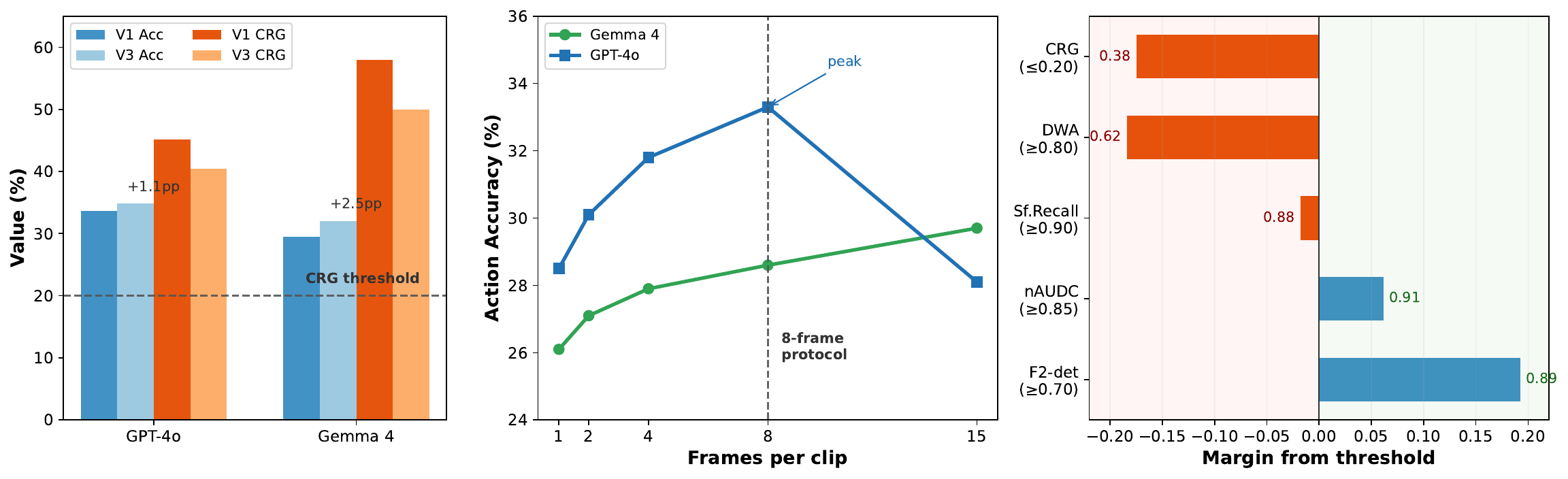}
    \caption{Protocol sensitivity analysis. (a) Prompt variants on a 150-clip
    stratified subset. (b) Frame-density analysis for GPT-4o and Gemma~4. (c)
    DRS threshold margins showing the gap between the best model value and
    each diagnostic threshold. Red bars indicate failing checks. Blue bars
    indicate passing checks.}
    \label{fig:ablations}
\end{figure}

We test whether the main findings are sensitive to prompt design, frame
sampling, ground-truth provenance, and evaluation thresholds.
Figure~\ref{fig:ablations} summarises the ablations.

\textbf{Prompt sensitivity.} We evaluate GPT-4o and Gemma~4 under three
prompt variants on 150 stratified clips. V1 uses the base protocol. V2
removes the taxonomy and asks for open-form descriptions. V3 adds structured
observation before taxonomy-based classification. Accuracy changes by at
most 2.5pp between V1 and V3, while default-action bias, low macro-F1, and
high CRG persist. V2 reduces B1 default frequency by 2 to 4 times, indicating
that part of the bias comes from taxonomy force-fitting. The persistence of
the bias under V3 suggests that prompting alone does not resolve the visual
ambiguity. Full metrics are in Appendix~\ref{app:prompt_sensitivity}.

\textbf{Frame density.} We run GPT-4o and Gemma~4 with 1, 2, 4, 8, and 15
frames per clip on the same subset. Gemma~4 improves from 26.1\% to 29.7\%,
while GPT-4o peaks at 8 frames and declines at 15 frames. The 8-frame protocol balances information density and model capacity. For these two
models, more frames do not close the human-model gap, suggesting that
surveillance-distance action discrimination is the main bottleneck. Full
curves are in Appendix~\ref{app:frame_density}.

\textbf{Threshold robustness.} The margin chart in
Figure~\ref{fig:ablations}(c) shows that most diagnostic failures are not
borderline. The best CRG (0.375) is nearly twice the 0.20 threshold, and
the best DWA (0.616) falls well short of 0.80. Only safety recall is close
to its threshold (best 0.882 versus 0.90). CRG at 0.20 tolerates a 20\% safety reasoning
failure rate even on correctly recognised actions. DWA at 0.80 requires that
cross-group confusions remain infrequent. Relaxing all thresholds by 20\%
would not change which checks pass for DWA or CRG.

\section{Related Work}
\label{sec:related}

\textbf{Industrial Action Recognition and Model Evaluation:}
Action recognition benchmarks such as Kinetics~\citep{kinetics}, ActivityNet~\citep{activitynet}, AVA~\citep{ava}, Ego4D~\citep{ego4d}, Assembly101~\citep{assembly101}, and MECCANO~\citep{meccano2021} have progressively moved toward realistic activity understanding, while Toyota SmartHome~\citep{toyotasmarthome} and anomaly-detection datasets~\citep{ucfcrime, shanghaitech} introduced surveillance viewpoints. Recent industrial benchmarks such as IndustryEQA~\citep{industryeqa} and iSafetyBench~\citep{isafetybench} rely on simulated or publicly available videos, whereas InspectSafe v1~\citep{inspecsafe} uses data from industrial robots. Existing safety datasets mainly focus on PPE detection without behaviour analysis~\citep{constructionsafetyai, vlmsafetyinterpretable}. General VLM benchmarks, including MMMU~\citep{mmmu}, Video-MME~\citep{videomme}, MVBench~\citep{mvbench}, ImageNet-C~\citep{imagenetc}, and NaturalBench~\citep{naturalbench}, also do not capture the complexity of industrial surveillance. Our prior work~\citep{actionsafety} explored industrial safety analysis using object-detection-based action priors. 


\textbf{Annotation Reliability Audit:}
At surveillance distance, visual evidence is often ambiguous enough that
annotators benefit from model-generated pre-fills, but those pre-fills can
also influence the resulting labels. \citet{schroeder2025humanloop} show that
LLM-generated suggestions shift human labelling behaviour, particularly on
ambiguous tasks, and \citet{wang2024humanllm} studies how to verify
LLM-generated labels effectively. \citet{preferenceleakage2025} demonstrates
that same-family contamination biases model-based evaluation. An assessment
of dataset curation practices across NeurIPS Datasets \& Benchmarks submissions found that benchmark papers rarely measure whether model pre-fills affect final labels or downstream evaluation
outcomes~\citep{datacuration2024}. 




\section{Conclusion}
\label{sec:conclusion}

We introduced SteelBench, a diagnostic benchmark for evaluating VLMs on
industrial activity recognition, safety reasoning, and annotation
provenance. The benchmark contains 1,345 clips annotated through a
four-stage pipeline with provenance tracked at every stage. Evaluating
9 cross-family VLMs, including frontier and large open weight, we find that different capabilities fail independently across models. The audit results further show that evaluation outcomes depend strongly on annotation provenance, reinforcing the need for provenance-stratified reporting in model-assisted benchmarks.

\textbf{Limitations:} SteelBench is scoped to a single integrated steel plant that houses 6 functionally distinct production domains. We found that cross-domain accuracy varies 8-16pp per model, but cross-facility generalisation remains untested. The three-tier review pipeline reduces harmful anchoring to 3.9\%, but does not eliminate it. The 8-frame representation captures temporal snapshots rather than
continuous video, and temporal reasoning tasks, such as action duration estimation, are not supported.

\textbf{Ethics and Data Release:} All footage was recorded by the existing plant security infrastructure with institutional approval from plant management, and faces are Gaussian-blurred to prevent individual identification. The dataset is released under CC-BY-NC-4.0 on HuggingFace, and the codebase is available on GitHub.

\textbf{Future work:} We plan to expand SteelBench, and we plan to at least double the clip count in the next release cycle. Raw footage from additional industrial domains, including power generation and mining, has already been collected for this purpose. We will also open-source our custom-built annotation tool with its built-in provenance audit module, so that the community can construct model-assisted benchmarks with integrated quality tracking.

\newpage
\begin{ack}
\end{ack}


\medskip

{\small
\bibliographystyle{unsrtnat}
\bibliography{References/references}

\begin{thebibliography}{28}
\providecommand{\natexlab}[1]{#1}
\providecommand{\url}[1]{\texttt{#1}}
\expandafter\ifx\csname urlstyle\endcsname\relax
  \providecommand{\doi}[1]{doi: #1}\else
  \providecommand{\doi}{doi: \begingroup \urlstyle{rm}\Url}\fi

\bibitem[Carreira and Zisserman(2017)]{kinetics}
Jo{\~a}o Carreira and Andrew Zisserman.
\newblock Quo vadis, action recognition? a new model and the kinetics dataset.
\newblock In \emph{CVPR}, 2017.

\bibitem[Heilbron et~al.(2015)Heilbron, Escorcia, Ghanem, and Niebles]{activitynet}
Fabian~Caba Heilbron, Victor Escorcia, Bernard Ghanem, and Juan~Carlos Niebles.
\newblock {ActivityNet}: A large-scale video benchmark for human activity understanding.
\newblock In \emph{CVPR}, 2015.

\bibitem[Gu et~al.(2018)Gu, Sun, Ross, Vondrick, Pantofaru, Li, Vijayanarasimhan, Toderici, Ricco, Sukthankar, et~al.]{ava}
Chunhui Gu, Chen Sun, David~A Ross, Carl Vondrick, Caroline Pantofaru, Yeqing Li, Sudheendra Vijayanarasimhan, George Toderici, Susanna Ricco, Rahul Sukthankar, et~al.
\newblock {AVA}: A video dataset of spatio-temporally localized atomic visual actions.
\newblock In \emph{CVPR}, 2018.

\bibitem[Grauman et~al.(2022)]{ego4d}
Kristen Grauman et~al.
\newblock Ego4d: Around the world in 3,000 hours of egocentric video.
\newblock In \emph{CVPR}, 2022.

\bibitem[Sener et~al.(2022)Sener, Chatterjee, Sheber, et~al.]{assembly101}
Fadime Sener, Dipika Chatterjee, Daniel Sheber, et~al.
\newblock Assembly101: A large-scale multi-view video dataset for understanding procedural activities.
\newblock In \emph{CVPR}, 2022.

\bibitem[Li et~al.(2025)Li, Chen, Dao, Li, Cai, Tan, Chen, and Kong]{industryeqa}
Yifan Li, Yuhang Chen, Anh Dao, Lichi Li, Zhongyi Cai, Zhen Tan, Tianlong Chen, and Yu~Kong.
\newblock {IndustryEQA}: Pushing the frontiers of embodied question answering in industrial scenarios.
\newblock In \emph{NeurIPS Datasets and Benchmarks}, 2025.

\bibitem[Ahmad et~al.(2024)]{sh17dataset}
Hafiz~Mughees Ahmad et~al.
\newblock {SH17}: A dataset for human safety and personal protective equipment detection in manufacturing industry.
\newblock \emph{Journal of Safety Science and Resilience}, 2024.

\bibitem[Guo et~al.(2025)Guo, Wong, Cheng, Chan, Leung, and Tao]{constructionsafetyai}
Koi~Xiaowen Guo, Peter Kok-Yiu Wong, Jack C.~P. Cheng, Chak-Fu Chan, Pak~Him Leung, and Xingyu Tao.
\newblock Enhancing visual-llm for construction site safety compliance via prompt engineering and bi-stage retrieval-augmented generation.
\newblock \emph{Automation in Construction}, 2025.
\newblock URL \url{https://api.semanticscholar.org/CorpusID:281021165}.

\bibitem[Chen et~al.(2024)Chen, Chen, Imani, Chen, and Imani]{vlmsafetyinterpretable}
Zhiling Chen, Hanning Chen, Mohsen Imani, Ruimin Chen, and Farhad Imani.
\newblock Vision language model for interpretable and fine-grained detection of safety compliance in diverse workplaces.
\newblock \emph{Expert Systems with Applications}, 265:\penalty0 125769, 11 2024.
\newblock \doi{10.1016/j.eswa.2024.125769}.

\bibitem[Wang et~al.(2024)Wang, Kim, Rahman, Mitra, and Miao]{wang2024humanllm}
Xinru Wang, Hannah Kim, Sajjadur Rahman, Kushan Mitra, and Zhengjie Miao.
\newblock Human-{LLM} collaborative annotation through effective verification of {LLM} labels.
\newblock In \emph{Proceedings of the 2024 CHI Conference on Human Factors in Computing Systems}. ACM, 2024.
\newblock \doi{10.1145/3613904.3641960}.

\bibitem[Akhtar et~al.(2024)]{datacuration2024}
Eshta Akhtar et~al.
\newblock The state of data curation at {NeurIPS}: An assessment of dataset development practices in the datasets and benchmarks track.
\newblock In \emph{Advances in Neural Information Processing Systems (NeurIPS), Datasets and Benchmarks Track}, 2024.
\newblock arXiv:2410.22473.

\bibitem[Kay et~al.(2017)Kay, Carreira, Simonyan, Zhang, Hillier, Vijayanarasimhan, Viola, Green, Back, Natsev, Suleyman, and Zisserman]{Kay2017TheKH}
Will Kay, Jo{\~a}o Carreira, Karen Simonyan, Brian~Hu Zhang, Chloe Hillier, Sudheendra Vijayanarasimhan, Fabio Viola, Tim Green, Trevor Back, Apostol Natsev, Mustafa Suleyman, and Andrew Zisserman.
\newblock The kinetics human action video dataset.
\newblock \emph{ArXiv}, abs/1705.06950, 2017.
\newblock URL \url{https://api.semanticscholar.org/CorpusID:27300853}.

\bibitem[Soomro et~al.(2012)Soomro, Zamir, and Shah]{Soomro2012UCF101AD}
Khurram Soomro, Amir Zamir, and Mubarak Shah.
\newblock Ucf101: A dataset of 101 human actions classes from videos in the wild.
\newblock \emph{ArXiv}, abs/1212.0402, 2012.
\newblock URL \url{https://api.semanticscholar.org/CorpusID:7197134}.

\bibitem[Wong et~al.(2026)]{preferenceleakage2025}
Howie Wong et~al.
\newblock Preference leakage: A contamination problem in {LLM}-as-a-judge.
\newblock In \emph{International Conference on Learning Representations (ICLR)}, 2026.
\newblock arXiv:2502.01534.

\bibitem[Schroeder et~al.(2025)Schroeder, Roy, and Kabbara]{schroeder2025humanloop}
Nico Schroeder, Subhajit Roy, and Jad Kabbara.
\newblock Just put a human in the loop? investigating {LLM}-assisted annotation for subjective tasks.
\newblock In \emph{Findings of the Association for Computational Linguistics (ACL)}, 2025.

\bibitem[Gebru et~al.(2021)Gebru, Morgenstern, Vecchione, Vaughan, Wallach, III, and Crawford]{gebru2021datasheets}
Timnit Gebru, Jamie Morgenstern, Briana Vecchione, Jennifer~Wortman Vaughan, Hanna Wallach, Hal~Daum{\'e} III, and Kate Crawford.
\newblock Datasheets for datasets.
\newblock \emph{Communications of the ACM}, 64\penalty0 (12):\penalty0 86--92, 2021.

\bibitem[Ragusa et~al.(2021)Ragusa, Furnari, Livatino, and Farinella]{meccano2021}
Francesco Ragusa, Antonino Furnari, Salvatore Livatino, and Giovanni~Maria Farinella.
\newblock The {MECCANO} dataset: Understanding human-object interactions from egocentric videos in an industrial-like domain.
\newblock In \emph{WACV}, 2021.

\bibitem[Das et~al.(2019)Das, Dai, Koperski, Minciullo, Garattoni, Bremond, and Francesca]{toyotasmarthome}
Srijan Das, Rui Dai, Michal Koperski, Luca Minciullo, Lorenzo Garattoni, Francois Bremond, and Gianpiero Francesca.
\newblock Toyota smarthome: Real-world activities of daily living.
\newblock In \emph{ICCV}, 2019.

\bibitem[Sultani et~al.(2018)Sultani, Chen, and Shah]{ucfcrime}
Waqas Sultani, Chen Chen, and Mubarak Shah.
\newblock Real-world anomaly detection in surveillance videos.
\newblock In \emph{CVPR}, 2018.

\bibitem[Liu et~al.(2018)Liu, Luo, Lian, and Gao]{shanghaitech}
Wen Liu, Weixin Luo, Dongze Lian, and Shenghua Gao.
\newblock Future frame prediction for anomaly detection --- a new baseline.
\newblock In \emph{CVPR}, 2018.

\bibitem[Abdullah et~al.(2025)Abdullah, Rawat, and Vyas]{isafetybench}
Raiyaan Abdullah, Yogesh~Singh Rawat, and Shruti Vyas.
\newblock {iSafetyBench}: A video-language benchmark for safety in industrial environments.
\newblock In \emph{IEEE/CVF International Conference on Computer Vision (ICCV) Workshops}, 2025.

\bibitem[Liu et~al.(2026)Liu, Liu, Min, Zhang, Cen, Han, Hu, Meng, He, and Zhou]{inspecsafe}
Zeyi Liu, Shuang Liu, Jihai Min, Zhaoheng Zhang, Jun Cen, Pengyu Han, Songqiao Hu, Zihan Meng, Xiao He, and Donghua Zhou.
\newblock Inspecsafe-v1: A multimodal benchmark for safety assessment in industrial inspection scenarios, 2026.

\bibitem[Yue et~al.(2024)]{mmmu}
Xiang Yue et~al.
\newblock {MMMU}: A massive multi-discipline multimodal understanding and reasoning benchmark for expert {AGI}.
\newblock In \emph{CVPR}, 2024.

\bibitem[Fu et~al.(2025)]{videomme}
Chaoyou Fu et~al.
\newblock Video-{MME}: The first-ever comprehensive evaluation benchmark of multi-modal {LLM}s in video analysis.
\newblock In \emph{CVPR}, 2025.

\bibitem[Li et~al.(2024{\natexlab{a}})]{mvbench}
Kunchang Li et~al.
\newblock {MVBench}: A comprehensive multi-modal video understanding benchmark.
\newblock In \emph{CVPR}, 2024{\natexlab{a}}.

\bibitem[Hendrycks and Dietterich(2019)]{imagenetc}
Dan Hendrycks and Thomas Dietterich.
\newblock Benchmarking neural network robustness to common corruptions and perturbations.
\newblock In \emph{ICLR}, 2019.

\bibitem[Li et~al.(2024{\natexlab{b}})]{naturalbench}
Baiqi Li et~al.
\newblock {NaturalBench}: Evaluating vision-language models on natural adversarial samples.
\newblock In \emph{NeurIPS Datasets and Benchmarks}, 2024{\natexlab{b}}.

\bibitem[Yarrabothula et~al.(2025)Yarrabothula, Kurrey, Nagar, and Gupta]{actionsafety}
Suryanarayana~Reddy Yarrabothula, Vaibhav Kurrey, Mayank Nagar, and Gagan~Raj Gupta.
\newblock Industrial safety violation detection using action recognition.
\newblock In \emph{Proceedings of the 8th International Conference on Data Science and Management of Data (12th ACM IKDD CODS and 30th COMAD)}, 2025.

\end{thebibliography}
}

\clearpage

\appendix
\section*{Appendix}
\section{Dataset Construction Details}
\label{app:pipeline}

Additional pipeline details are provided here, complementing the dataset description in Section~\ref{sec:dataset}. Distribution statistics are shown in Figure~\ref{fig:distributions} of the main paper.

\subsection{Curation Pipeline}

\textbf{Person detection.} YOLOv8n scans each video at 1 fps, triggering clip extraction when at least one person is detected with confidence above 0.25 sustained for 5 or more consecutive frames. This threshold is deliberately permissive. We prefer including ambiguous clips over missing genuine activity, because borderline cases (partially occluded workers, distant figures in steam) are precisely the conditions that test VLM perception. This stage produces 10,024 candidate clips from 117 source videos.

\textbf{Frame representation.} Each clip is represented by 8 frames sampled at fixed proportional positions (0\%, 14\%, 28\%, 43\%, 57\%, 71\%, 86\%, 100\% of clip duration) at original 1080p resolution. The 8-frame protocol balances temporal coverage with API cost and model context constraints. Our frame-density ablation (Appendix~\ref{app:frame_density}) confirms that accuracy plateaus near 8 frames for most models.

\textbf{Temporal deduplication}. Fixed cameras produce highly correlated consecutive clips. We group consecutive clips from the same camera within 60-second windows and retain at most 4 per group, prioritising visual diversity. This reduces the pool from 10,024 to approximately 5,000 temporally unique clips.

\textbf{Class-balanced curation.} From the deduplicated pool, we apply stratified sampling across action class, site, visibility condition, and scene type to produce a 3,304-clip annotation batch. The target is a minimum of 15 clips per action class, where available.

\textbf{Annotation and verification.} The four-stage annotation pipeline (Section~\ref{sec:audit}) processes the batch, producing 1,345 clips that pass quality validation. Clips flagged, abandoned, or that failed schema validation are excluded.

\textbf{Cost.} Total marginal cost: \$71 (VLM pre-annotation \$21 via API, infrastructure and compute \$50). No dedicated GPU was required for annotation.

\subsection{Full Action Taxonomy}

Table~\ref{tab:app_taxonomy} lists all 25 evaluated action classes with group, description, and instance count.

\begin{table}[h]
\centering
\caption{Full 25-class action taxonomy with instance counts from the final 1,345-clip evaluation set.}
\label{tab:app_taxonomy}
\scriptsize
\begin{tabular}{llllr}
\toprule
\textbf{Code} & \textbf{Group} & \textbf{Name} & \textbf{Description} & \textbf{$n$} \\
\midrule
A1 & A Locomotion & Walking & Moving on foot & 808 \\
A2 & A Locomotion & Climbing up & Ascending stairs/ladder & 42 \\
A3 & A Locomotion & Climbing down & Descending stairs/ladder & 38 \\
A4 & A Locomotion & Carrying/walking & Walking while carrying an object & 154 \\
A5 & A Locomotion & Pushing/pulling & Moving objects by pushing/pulling & 73 \\
\midrule
B1 & B Stationary Work & Operating (tool use) & Standing operation with tools & 765 \\
B2 & B Stationary Work & Panel operation & Operating control panels & 125 \\
B3 & B Stationary Work & Welding & Hot work with welding equipment & 52 \\
B4 & B Stationary Work & Hot work (non-weld) & Hot work without welding & 20 \\
B5 & B Stationary Work & Floor work & Crouching or kneeling work & 564 \\
B6 & B Stationary Work & Overhead reaching & Working above head height & 31 \\
B7 & B Stationary Work & Inspection & Stationary visual inspection & 578 \\
B8 & B Stationary Work & Lifting/placing & Manual lift and place & 82 \\
B9 & B Stationary Work & Sitting operation & Seated equipment operation & 156 \\
\midrule
C1 & C Crane/Equip & Crane signaling & Hand signals to crane operator & 26 \\
C2 & C Crane/Equip & Crane hook interaction & Attaching/detaching hooks & 36 \\
C3 & C Crane/Equip & Guiding load & Guiding suspended load & 33 \\
C4 & C Crane/Equip & Vehicle operating & Driving industrial vehicle & 9$^\dagger$ \\
\midrule
D1 & D Material Hand. & Team carry & Coordinated manual carrying & 66 \\
D2 & D Material Hand. & Loading/unloading & Loading or unloading materials & 65 \\
D3 & D Material Hand. & Cylinder handling & Moving gas cylinders & 4$^\dagger$ \\
\midrule
E1 & E Social & Communicating & Verbal or gestural communication & 414 \\
E2 & E Social & Supervising & Overseeing others' work & 296 \\
\midrule
F1 & F Idle & Idle standing & Standing without activity & 838 \\
F2 & F Idle & Idle sitting & Sitting without activity & 304 \\
\bottomrule
\end{tabular}
\end{table}

 \subsection{Annotation Scale Comparison}
  \label{app:scale_comparison}

  Table~\ref{tab:app_scale_comparison} compares SteelBench against
  established video benchmarks on annotation density. Standard action
  recognition datasets provide a single label per clip. SteelBench
  annotates 9 to 58 structured fields per clip, yielding comparable total annotation volume to datasets with 10 to 100$\times$ more
  clips.

  \begin{table}[h]
  \centering
  \caption{Annotation scale comparison. ``Fields/clip'' counts distinct
  annotation decisions per instance. SteelBench compensates for fewer
  clips with substantially denser per-clip annotation.}
  \label{tab:app_scale_comparison}
  \small
  \begin{tabular}{lrrrl}
  \toprule
  \textbf{Benchmark} & \textbf{Clips} & \textbf{Fields/clip} & \textbf{Total decisions} & \textbf{Domain} \\
  \midrule
  Kinetics-400 & 300,000 & 1 & 300,000 & Internet video \\
  ActivityNet v1.3 & 19,994 & 1--2 & ${\sim}$23,000 & Internet video \\
  UCF101 & 13,320 & 1 & 13,320 & Internet video \\
  Toyota Smarthome & 16,000 & 1 & 16,000 & Indoor ADL \\
  HAA500 & 10,000 & 1 & 10,000 & Internet video \\
  IndustryEQA & 76 videos & ${\sim}$18 & ${\sim}$1,344 & Simulated warehouse \\
  \midrule
  \textbf{SteelBench} & \textbf{1,345} & \textbf{${\sim}$ 9 - 58(mean ~24)} & \textbf{${\sim}$~42,000} & \textbf{Real industrial CCTV} \\
  \bottomrule
  \end{tabular}
  \end{table}

Each clip carries up to structured annotation fields, depending on the annotation layer and worker count. Layer 1 clips (scene-level, 538 clips) require 9 fields. Layer 2 clips (per-person, 807 clips) require 8 scene-level fields plus 14 fields per worker, yielding 9-58 fields per clip. Across the full dataset, this yields ~42,000 individual annotation decisions, including optional fields such as free-text descriptions, physical descriptions, and confidence levels.

\subsection{Safety Rule Book Summary}

The facility safety rule book contains 55 general rules (UA-G-01 through UA-G-55) and site-specific rules for 8 departments. In practice, 18 general rules and 7 site-specific rules are cited in the GT annotations. General rules cover PPE compliance (helmets, harnesses, safety shoes, welding protection), safe work practices (proper lifting posture, use of guide ropes during crane operations), and prohibited behaviours (mobile phone use near machinery, working at height without a harness). Department-specific rules address zone-specific hazards such as fire-resistant suits near blast furnaces (UA-BF-03), dust masks in sinter plant areas (UA-SP-01), and vehicle movement restrictions in rolling mill shops (UA-CRM-01). The complete rule book is included in the dataset release.

\subsection{Example Annotation}

The following JSON excerpt shows a representative Layer 2 annotation for clip \texttt{clip\_RERS\_RERS-1\_20251118\_000655\_0025} (4 workers, dust and glare, MAI scene). Two of four workers are shown for space:

\begin{lstlisting}[basicstyle=\ttfamily\scriptsize, breaklines=true]
{
  "clip_id": "clip_RERS_RERS-1_20251118_000655_0025",
  "annotation_layer": 2,
  "scene_type": "MAI",
  "num_workers": 4,
  "visibility_conditions": ["dust", "glare"],
  "persons": [
    {
      "person_id": "P1",
      "action_code": "B2",
      "position": "left",
      "physical_description": "worker in blue coveralls and
                               yellow helmet, stocky build",
      "ppe": {"helmet": "worn", "high_vis_vest": "worn",
              "safety_shoes": "worn",
              "welding_protection": "not_applicable",
              "harness": "not_applicable"},
      "spatial_context": ["ground_level", "near_machinery"],
      "unsafe_act": "none",
      "group_flag": "solo"
    },
    {
      "person_id": "P3",
      "action_code": "B1",
      "position": "centre",
      "physical_description": "worker in blue coveralls and
                               yellow helmet, on platform",
      "ppe": {"helmet": "worn", "high_vis_vest": "not_worn",
              "safety_shoes": "cannot_determine",
              "harness": "not_worn"},
      "spatial_context": ["elevated", "near_machinery",
                          "crane_zone"],
      "unsafe_act": "working at height without harness",
      "group_flag": "solo"
    }
  ]
}
\end{lstlisting}

Worker P3 triggers a safety violation (UA-G-35: working at height with improper harness anchoring). The annotation captures the action (B1 operating), the missing PPE (harness not worn), the spatial context (elevated, crane zone), and the specific unsafe act. This illustrates the structured per-worker annotation density provided by SteelBench.

\section{Annotation Quality Details}
\label{app:audit_details}

This section provides the complete annotation quality analyses supporting Section~\ref{sec:audit}. Direction analyses use expert-verified ground truth as the reference standard. IAA is computed on 289 double-annotated pairs (102 blind, 192 anchored).

\subsection{Per-Field Direction Analysis}

\begin{figure}[h]
    \centering
    \includegraphics[width=\textwidth]{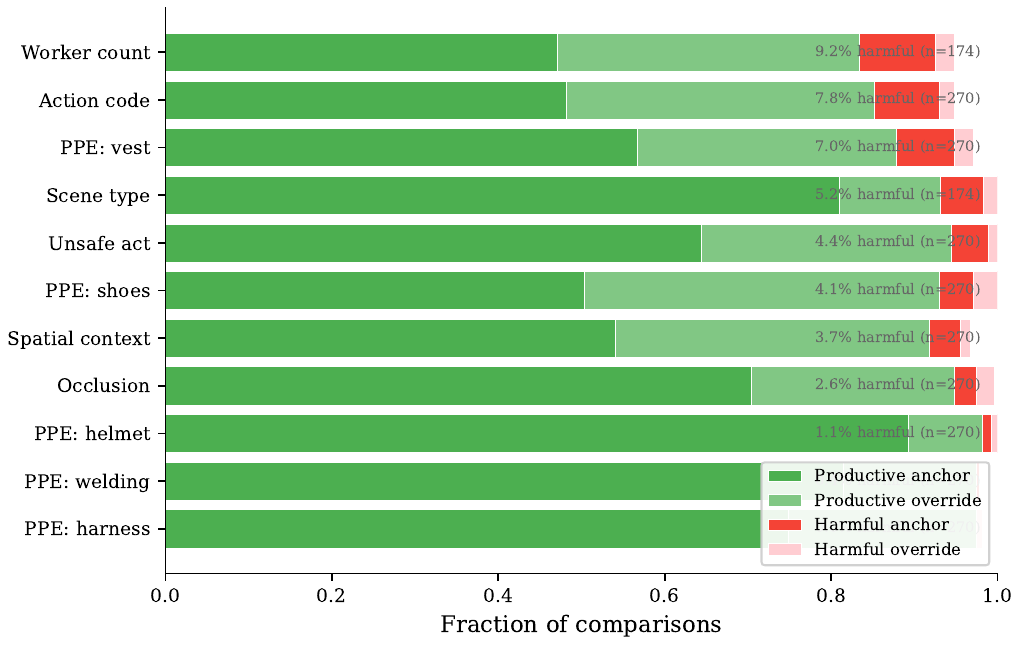}
    \caption{Direction analysis by annotation field (2,829 comparisons across 177 proper-chain clips). Harmful anchoring ranges from 0\% (harness) to 9.0\% (worker count), following a gradient of visual verifiability at 7--10m camera distance.}
    \label{fig:app_direction_field}
\end{figure}

The main paper reports an aggregate 92.8\% correct label rate across all annotation fields (Section~\ref{sec:audit}). Figure~\ref{fig:app_direction_field} decomposes this by individual field, revealing that harmful anchoring is not uniformly distributed --- it follows a gradient determined by the visual verifiability of each field at surveillance distance.

Fields cluster into three regimes. \textbf{Binary PPE items} visible from overhead --- helmets (1.1\% harmful anchoring), harnesses (0\%) --- can be independently verified by the annotator, and harmful anchoring is near-zero. \textbf{Categorical fields} requiring fine-grained discrimination --- action code (8.0\%), worker count (9.0\%) --- involve distinguishing between visually similar states at distances where the discriminating cues (hand position, tool presence, partial occlusion) occupy fewer than 10 pixels, and harmful anchoring peaks. \textbf{Spatial context} (3.6\% harmful anchoring) presents a distinct pattern: it has the highest anchoring \emph{bias} of any field (+44pp blind-to-anchored difference in VLM agreement), yet relatively low harmful anchoring. The dissociation indicates that annotators frequently defer to VLM spatial tags, and the VLM is usually correct on spatial layout --- the deference inflates bias but does not proportionally inflate error.

This gradient has a practical implication: harmful anchoring concentrates in exactly the fields where annotation is genuinely difficult, not where annotators are careless. The same visual ambiguities that produce harmful anchors also produce the within-group IAA confusions analysed in Appendix~\ref{app:iaa_confusion}.

\subsection{Per-Annotator Direction Analysis}
\label{app:annotator_bias}

\begin{figure}[h]
    \centering
    \includegraphics[width=\textwidth]{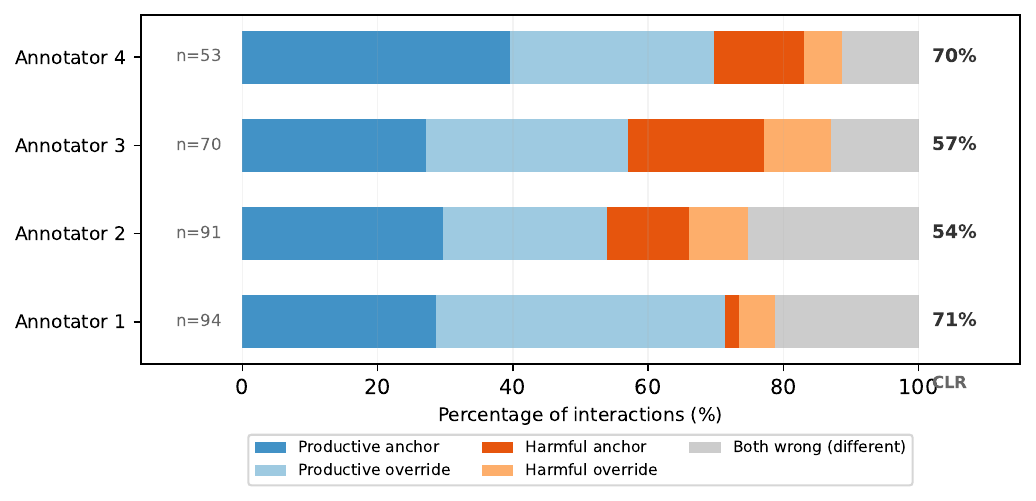}
    \caption{Per-annotator direction analysis on action classification.
    Blue segments are productive outcomes (VLM helped or the annotator corrected).
    Red and orange segments are harmful outcomes. Grey indicates cases where
    Both VLM and the annotator were wrong with different errors. CLR is the
    correct-label rate (productive anchor + productive override). These are
    action-level comparisons on the expert-overlapping subset and differ in
    denominator from the aggregate field-level CLR of 92.8\% reported in
    Section~\ref{sec:audit_findings}.}
    \label{fig:app_annotator_direction}
\end{figure}

Figure~\ref{fig:app_annotator_direction} decomposes the direction analysis
by annotator, showing how each annotator interacts with VLM pre-fills on
action classification. The analysis covers clips where VLM pre-fill,
annotator label, and expert reference are all available.

Annotator~1 has the highest correct-label rate (71\%) and the lowest harmful
anchoring (2.1\%), indicating strong independent judgment. Annotator~4 shows
the highest productive anchoring (39.6\%), meaning they agree with the VLM
more often when it is correct, while maintaining low harmful anchoring
(13.2\%). Annotator~3 has the highest harmful anchor rate (20.0\%) on this
subset, though the sample is small ($n$=70) and their overall edit depth
remains high (10.9 fields per clip, Table~\ref{tab:app_annotator_perf}).
The variation across annotators shows that VLM influence is not uniform.
Different annotators adopt different strategies for deciding when to defer
to the VLM and when to override it. The three-tier verification pipeline
ensures that harmful anchors from any individual annotator are caught by
expert review downstream.

\subsection{Annotator Engagement}
Two patterns are notable. First, annotation time is substantial regardless of whether the VLM action code is retained (mean 446s) or overridden (527s) - the 81s difference indicates that retaining a VLM suggestion still involves reviewing ${\sim}$15 other fields rather than accepting the pre-fill wholesale. Second, flagging rates vary from 4\% (annotator~3) to 9\% (annotator~1), and annotator~1's high flagging rate coincides with the highest unsafe-act detection rate (36\%), indicating that flagging reflects active safety evaluation rather than avoidance of difficult clips.


\begin{table*}[t]
\centering
\caption{Per-annotator statistics. Edit depth denotes the mean number of modified fields per clip (out of ${\sim}$15 editable). Zero-mod \% indicates submissions without changes from the VLM pre-fill. Annotator~10 was identified by the audit protocol and excluded from GT construction (Appendix~\ref{app:annotator10_case}).}
\label{tab:app_annotator_perf}
\small
\setlength{\tabcolsep}{5pt}
\begin{tabular}{lccccccc}
\toprule
\textbf{Annotator} & \textbf{Subm.} & \textbf{Flagged} & \textbf{Conf.} & \textbf{Edit} & \textbf{Zero-mod} & \textbf{Unsafe} & \textbf{Median} \\
 &  &  &  & \textbf{depth} & \textbf{(\%)} & \textbf{(\%)} & \textbf{time} \\
\midrule
Annotator 1 & 434 & 38 & 0.883 & 10.2 & 1\%  & 36\% & 5.2 m \\
Annotator 2 & 443 & 31 & 0.830 & 11.0 & 1\%  & 27\% & 2.9 m \\
Annotator 3 & 386 & 16 & 0.825 & 10.9 & 1\%  & 32\% & 4.3 m \\
Annotator 4 & 303 & 18 & 0.834 & 9.7  & 1\%  & 18\% & 3.8 m \\
\midrule
Annotator 10$^\dagger$ & 41 & 1 & 0.894 & 3.1 & 14\% & 38\% & 3.9 m \\
\bottomrule
\end{tabular}

\vspace{2pt}
\raggedright\scriptsize
$^\dagger$ Identified by the audit protocol and excluded from GT construction.
\end{table*}

\subsection{Annotation Quality Case Study: Annotator Exclusion}
\label{app:annotator10_case}

The provenance audit protocol (Section~\ref{sec:audit}) detected one annotator whose behaviour diverged from the rest of the team. Annotator\_10, a newly recruited Tier-1 worker, submitted 41 clips over four days. Table~\ref{tab:annotator10} compares their quality indicators against the four retained annotators.

  \begin{table}[h]
  \centering
  \caption{Quality indicators for the excluded annotator vs.\ the four
  retained Tier-1 annotators. All metrics indicate that annotator\_10
  accepted VLM pre-fills with minimal review.}
  \label{tab:annotator10}
  \small
  \begin{tabular}{lcc}
  \toprule
  \textbf{Metric} & \textbf{Annotator\_10} & \textbf{Retained (range)} \\
  \midrule
  Clips submitted         & 41        & 318--454 \\
  Action override rate     & 16.2\%   & 36--53\% \\
  PPE override rate        & 24.3\%   & 67--92\% \\
  $\kappa$ vs.\ VLM       & 0.775    & 0.43--0.59 \\
  VLM agreement rate       & 82.4\%   & 51--64\% \\
  Avg.\ confidence         & 0.894    & 0.71--0.85 \\
  Productive overrides     & 0        & 11--35 \\
  Harmful anchor rate      & 17\%     & 4--15\% \\
  \bottomrule
  \end{tabular}
  \end{table}

Three signals triggered exclusion. First, the override rate across all fields was 2--4$\times$ lower than the team range, suggesting the annotator accepted VLM outputs without substantive review. Second, VLM agreement ($\kappa$=0.775) was far above the team ceiling of 0.59, consistent with anchoring rather than independent judgment. Third, direction analysis found zero productive overrides across 6 evaluated interactions. The annotator never corrected a VLM error.

Excluding these 41 clips reduced the ground truth from approximately 1,370 to 1,345 clips. Every aggregate quality metric improved. Override rate increased from 94.1\% to 97.0\%, harmful anchor rate dropped from 5.2\% to 3.9\%, and action $\kappa$ rose from 0.74 to 0.78. This case demonstrates that the audit protocol serves as a functional quality measuring tool.

\subsection{Expert Correction Rate}

Among 363 expert-reviewed clips, we identified 16 cases of harmful anchoring: instances in which the VLM was incorrect, the Tier-1 annotator accepted the error, and the expert subsequently corrected it. Experts corrected all 16, achieving a 100

For the remaining 982 GT clips (73\%) finalised at Tier~1 without expert review, the 3.9\% full-field harmful anchor rate yields an upper bound of ${\sim}$38 clips with potentially uncorrected harmful anchors (2.8\% of the evaluation set). This is a worst-case estimate: it assumes no self-correction by Tier-1 annotators and treats all harmful anchors equally, although many involve secondary fields (spatial tags, occlusion levels) rather than primary evaluation targets (action codes, PPE states).

\subsection{Safety-Specific Direction Analysis}

Safety violations follow a distinct direction pattern from other annotation fields. Table~\ref{tab:app_safety_direction} decomposes 169 person-level safety comparisons.

\begin{table}[h]
\centering
\caption{Safety-specific direction analysis ($n$=169 person-level comparisons). The VLM's dominant error mode is over-flagging (34.3\%), while annotators independently identify violations the VLM missed (8.9\%).}
\label{tab:app_safety_direction}
\small
\begin{tabular}{lcc}
\toprule
\textbf{Outcome} & \textbf{Count} & \textbf{\%} \\
\midrule
VLM false alarm, annotator corrected & 58 & 34.3 \\
Both agree unsafe & 51 & 30.2 \\
Both agree safe & 38 & 22.5 \\
Annotator found violation VLM missed & 15 & 8.9 \\
Both incorrect (shared failure) & 5 & 3.0 \\
Both missed real violation & 1 & 0.6 \\
\bottomrule
\end{tabular}
\end{table}

The asymmetry between VLM false alarms (58) and VLM missed violations (15) reflects the safety-officer persona used in the VLM prompt (Appendix~\ref{app:pipeline}), which biases the model toward over-flagging. From an annotation quality perspective, this is the preferable error direction: false alarms are corrected by annotators, while missed violations risk propagating to the GT uncaught. That only 1 of 169 comparisons (0.6\%) involved both annotator and VLM missing a real violation --- caught only by Tier-2 expert review --- indicates that the combination of VLM pre-fill and human review provides near-complete safety coverage within the reviewed subset.

\subsection{IAA Confusion Patterns}
\label{app:iaa_confusion}

Action agreement ($\kappa$=0.780, 289 DA pairs) is strong overall. Table~\ref{tab:app_iaa_confusion} identifies where disagreements concentrate.

\begin{table}[h]
\centering
\caption{Top within-group action disagreements (annotators agree on group but differ on class). Group~B (stationary work, 9 classes) accounts for 14 of 22 within-group confusions.}
\label{tab:app_iaa_confusion}
\small
\begin{tabular}{lcc}
\toprule
\textbf{Confusion pair} & \textbf{Group} & \textbf{Count} \\
\midrule
B1 $\leftrightarrow$ B5 (operating $\leftrightarrow$ floor work) & B & 6 \\
B1 $\leftrightarrow$ B7 (operating $\leftrightarrow$ inspection) & B & 3 \\
B2 $\leftrightarrow$ B5 (panel work $\leftrightarrow$ floor work) & B & 3 \\
E1 $\leftrightarrow$ E2 (communicating $\leftrightarrow$ supervising) & E & 2 \\
B1 $\leftrightarrow$ B2 (operating $\leftrightarrow$ panel work) & B & 2 \\
F1 $\leftrightarrow$ F2 (standing idle $\leftrightarrow$ sitting idle) & F & 2 \\
\bottomrule
\end{tabular}
\end{table}

Group~B accounts for 64\% of within-group disagreements despite comprising 36\% of action classes (9/25), reflecting the visual similarity of stationary work activities from overhead cameras: operating machinery (B1), inspecting equipment (B7), and working at ground level (B5) are distinguished primarily by hand position and tool engagement, which are below reliable resolution at surveillance distance. This concentration has two implications. First, it identifies the specific taxonomy boundary where human annotation reaches its perceptual ceiling under overhead CCTV --- a ceiling that VLMs share, as evidenced by the B1-magnet bias in evaluation (Section~\ref{sec:results}). Second, it indicates that IAA disagreements are systematic rather than random, meaning that group-level accuracy (collapsing B-subclasses) would be substantially higher than the reported class-level $\kappa$.

\subsection{Blind Double-Annotation}

To measure whether VLM pre-fills inflate inter-annotator agreement, we constructed 102 blind double-annotated pairs where neither annotator saw VLM suggestions. Table~\ref{tab:app_blind_iaa} compares IAA across conditions.

\begin{table}[h]
\centering
\caption{IAA by annotation condition. Action and PPE agreement are stable across conditions. Spatial context shows a large drop under blind conditions ($-$0.213), identifying it as the field most influenced by VLM pre-fills.}
\label{tab:app_blind_iaa}
\small
\begin{tabular}{lcccc}
\toprule
\textbf{Condition} & \textbf{Action $\kappa$} & \textbf{PPE $\kappa$} & \textbf{Spatial $\kappa$} & \textbf{$n$ pairs} \\
\midrule
All DA pairs & 0.780 & 0.793 & 0.632 & 289 \\
Blind only & 0.766 & 0.805 & 0.419 & 102 \\
\bottomrule
\end{tabular}
\end{table}

The three fields exhibit distinct responses to removing VLM influence. Action $\kappa$ drops by only 0.014 (0.780 $\rightarrow$ 0.766), confirming that the reported action agreement reflects independent human consensus. PPE $\kappa$ \emph{increases} by 0.012 (0.793 $\rightarrow$ 0.805), indicating that VLM PPE suggestions occasionally introduce inconsistency --- annotators who would otherwise agree on a PPE state may second-guess their judgment when the VLM suggests otherwise. Spatial $\kappa$ drops by 0.213 (0.632 $\rightarrow$ 0.419), the largest shift, identifying spatial context as the field where VLM pre-fills have the strongest standardising effect. Without VLM vocabulary, annotators describe the same physical layout using inconsistent terminology (e.g., ``near conveyor belt'' vs ``beside rolling equipment'').

These three regimes align with the per-field direction analysis: fields where VLM influence on IAA is minimal (action, PPE) are the same fields that carry the primary evaluation weight, while the field most influenced (spatial context) serves as supplementary annotation metadata.

\subsection{GT Provenance Slice Definitions}                                                                                                                                                              
  \label{app:gt_slices}                                                                                                                                                                                   
                                                                   
  The provenance gradient (Section~\ref{sec:audit_findings}) evaluates the                                                                                                                                  
  annotation VLM (Qwen3-VL-235B) against three sets of reference labels.                                                                                                                                    
  All three evaluations use the same VLM predictions. Only the reference                                                                                                                                    
  labels differ. Table~\ref{tab:gt_slices} defines each slice.                                                                                                                                              
                                                                                                                                                                                                            
  \begin{table}[h]                                                                                                                                                                                          
  \centering                                                                                                                                                                                                
  \caption{Ground-truth provenance slices. Each slice defines a different                                                                                                                                   
  set of reference labels against which the same VLM predictions are                                                                                                                                        
  scored. Slices are not disjoint at the clip level because the same clip                                                                                                                                   
  may have annotations from multiple annotators under different                                                                                                                                             
  conditions. This does not bias the comparison because the VLM                                                                                                                                             
  predictions are fixed and only the reference changes.}                                                                                                                                                    
  \label{tab:gt_slices}                                                                                                                                                                                     
  \small                                                                                                                                                                                                    
  \begin{tabular}{lrp{6.5cm}}                                                                                                                                                                               
  \toprule                                                                                                                                                                                                  
  \textbf{Slice} & \textbf{$n$ (pairs)} & \textbf{Construction} \\                                                                                                                                        
  \midrule                                                                                                                                                                                                  
  Blind & 102 & Tier-1 annotators labeled these clips without seeing any               
  VLM suggestions. Labels reflect independent human judgment. \\                                                                                                                                            
  Proper-chain & 169 & Tier-1 annotators corrected VLM pre-fills, then                
  experts verified the Tier-1 annotations. Labels passed through VLM                                                                                                                                        
  $\rightarrow$ human $\rightarrow$ expert. \\                                                                                                                                                              
  VLM-sourced & 157 & Experts reviewed and edited VLM output directly                                                                                                                                       
  without a Tier-1 intermediate. Labels passed through VLM                                                                                                                                                  
  $\rightarrow$ expert. \\                                                                                                                                                                                  
  \bottomrule                                                                         
  \end{tabular}                                                                                                                                                                                             
  \end{table}                                                                         
                                                                                                                                                                                                            
  The gradient (37.2\% $\rightarrow$ 57.4\% $\rightarrow$ 77.7\%) measures
  how much the VLM's apparent accuracy changes as reference labels become                                                                                                                                   
  more influenced by the VLM itself. The blind slice provides the most                
  independent evaluation. The VLM-sourced slice has the strongest circular                                                                                                                                  
  dependency. The 41.1pp gap between blind and VLM-sourced accuracy quantifies the inflation that unaudited model-assisted annotation can 
  introduce.                                                                                                                                                                                                
                                                                                                                                                                                                            
  A clip may appear in multiple slices if it was annotated by both a                                                                                                                                        
  blind-condition annotator and an anchored-condition annotator. In such                                                                                                                                    
  cases, each annotator's labels enter their respective slice                                                                                                                                               
  independently. Because the VLM predictions are identical across all                                                                                                                                       
  three comparisons, clip-level overlap does not confound the measurement.            
  The comparison is not ``blind clips vs.\ anchored clips'' (which would
  require disjoint samples). It is ``same predictions, different                                                                                                                                            
  references'' (which does not).                                   
                                                                                  
\section{Metric Definitions}
\label{app:metrics}

This section provides formal metric definitions, evaluation protocol details,
and annotation guidelines. Section~\ref{sec:framework} introduces the
diagnostic metrics in the main text. Here we provide complete definitions
including DRS component metrics (DWA, safety recall, F2-detect), the full
prompt design, model configurations, output parsing rules, and annotator
instructions.

\subsection{Annotation Quality Metrics}

\subsubsection{Cohen's Kappa ($\kappa$)}

Measures inter-annotator agreement adjusted for chance:
\begin{equation}
\kappa = \frac{p_o - p_e}{1 - p_e}
\end{equation}
where $p_o$ is observed agreement and $p_e = \sum_c p_{1c} \cdot p_{2c}$ is expected chance agreement. Interpretation: $<$0.20 poor, 0.21--0.40 fair, 0.41--0.60 moderate, 0.61--0.80 substantial, 0.81--1.00 near-perfect.

We compute $\kappa$ on three axes across 289 DA pairs: action classification ($\kappa$=0.780, substantial), PPE assessment ($\kappa$=0.793, substantial), and spatial context ($\kappa$=0.632, substantial). Expert proper-chain $\kappa$=0.82 (near-perfect).

\subsubsection{Override Rate}

The fraction of clips where the annotator changed at least one field from VLM pre-fill:
\begin{equation}
\text{Override Rate} = \frac{|\{c : \exists\, f \in \text{fields},\; \text{human}(c,f) \neq \text{VLM}(c,f) \}|}{|\text{clips}|}
\end{equation}

A high rate indicates active curation. SteelBench: 97.0\% of 1,514 clips modified. Per-field rates range from 5.7\% (helmet) to 77.3\% (dominant actions).

\subsubsection{Direction Analysis}

Each annotator--VLM interaction is classified into four outcomes using expert ground truth as reference:

\begin{table}[htpb]
\centering
\caption{Direction analysis taxonomy for annotator--VLM interactions. Outcomes 
are categorized based on whether the VLM prediction matches expert ground truth 
and whether the annotator accepts or overrides the suggestion.}
\label{tab:direction_analysis}
\small
\begin{tabular}{llll}
\toprule
\textbf{VLM vs GT} & \textbf{Human action} & \textbf{Category} & \textbf{Desirable?} \\
\midrule
Correct & Accepts VLM & Productive anchor & \checkmark \\
Wrong & Corrects VLM & Productive override & \checkmark \\
Wrong & Accepts VLM & Harmful anchor & $\times$ \\
Correct & Changes VLM & Harmful override & $\times$ \\
\bottomrule
\end{tabular}
\end{table}

Derived metrics:
\begin{align}
\text{Correct label rate} &= \frac{\text{productive anchor} + \text{productive override}}{\text{total comparisons}} \\
\text{Correction-to-acceptance ratio} &= \frac{\text{productive override count}}{\text{harmful anchor count}}
\end{align}

The correct label rate measures whether VLM-assisted annotation produces accurate labels. The correction ratio quantifies how effectively annotators catch VLM errors relative to how often they accept them. SteelBench: 92.8\% correct labels, 6.9$\times$ correction ratio (2,829 field-level comparisons across 177 proper-chain clips). Per-field decomposition in Appendix~\ref{app:audit_details}.

\subsubsection{Anchoring Bias}

The difference in VLM agreement rate between anchored and blind annotation conditions:
\begin{equation}
\text{Anchoring Bias} = \text{Agreement}_{\text{anchored}} - \text{Agreement}_{\text{blind}}
\end{equation}

Positive bias indicates that seeing VLM suggestions shifts annotator behaviour toward the VLM's predictions. Anchoring bias measures \emph{influence}, not \emph{error} --- the relationship between bias and label quality depends on the direction of analysis. SteelBench: +13.9pp on action classification.

\subsubsection{ECE by GT Provenance}

Expected Calibration Error computed separately against ground truth constructed under three conditions:
\begin{itemize}[nosep]
    \item \textbf{Blind GT:} annotators labeled without seeing VLM suggestions
    \item \textbf{Proper-chain expert GT:} experts reviewed human annotations (not VLM output)
    \item \textbf{VLM-sourced expert GT:} experts reviewed VLM output directly
\end{itemize}

\begin{equation}                                                                                                                                                                                          
  \mathrm{ECE} = \sum_{b=1}^{B} \frac{n_b}{N}                                                                                                                                                               
  \left| \mathrm{acc}(b) - \mathrm{conf}(b) \right|                                   
  \end{equation}                                                                                                                                                                                            
  where $B$=10 equal-width bins, $n_b$ is the number of predictions in                
  bin $b$, $\mathrm{acc}(b)$ is the fraction correct in that bin, and $\mathrm{conf}(b)$ is the mean confidence. Confidence scores are
  self-reported by each VLM (not derived from log-probabilities).

If ECE differs substantially across conditions, the annotation process distorts calibration measurement. In SteelBench, the annotation VLM reports ${\sim}$90\% confidence in all conditions, but accuracy against blind GT (37.2\%, ECE=0.528) differs sharply from accuracy against VLM-sourced GT (77.7\%, ECE=0.117).

\subsection{Evaluation Metrics}

Standard accuracy and macro-F1 capture overall and class-balanced performance, respectively. The diagnostic metrics each target a specific failure mode relevant to industrial surveillance: nAUDC tests robustness to environmental degradation, CRG tests whether perception translates into reasoning, safety recall tests violation-detection coverage, DWA tests whether errors are taxonomically severe, and F2-detect tests basic worker detection. DRS aggregates these into a diagnostic checklist.

\subsubsection{Weighted Action Accuracy}

\begin{equation}
\text{Accuracy} = \frac{1}{N}\sum_{i=1}^{N} \mathbb{1}[\hat{a}_i = a_i]
\end{equation}
where $\hat{a}_i$ and $a_i$ are predicted and GT action codes for person $i$. Measures how often the model correctly identifies worker activity.

\subsubsection{Macro-F1}

\begin{equation}
\text{Macro-F1} = \frac{1}{|C|} \sum_{c \in C} \frac{2 \cdot P_c \cdot R_c}{P_c + R_c}
\end{equation}
where $P_c$ and $R_c$ are per-class precision and recall. Equal weight to all classes regardless of frequency, exposing failures on rare safety-critical actions that weighted accuracy conceals.

\subsubsection{nAUDC (Normalised Area Under Degradation Curve)}

\begin{equation}
\text{AUDC} = \frac{1}{|V|} \sum_{v \in V} \text{Acc}_v, \qquad \text{nAUDC} = \frac{\text{AUDC}}{\max_v \text{Acc}_v}
\end{equation}
where $V = \{\text{clear, dust, glare, low\_light, steam, smoke}\}$. Multi-condition clips contribute to all applicable bins.

\textbf{Rationale.} Standard accuracy on clear-condition subsets overestimates deployed performance. Industrial environments cause persistent visibility degradation that better cameras cannot mitigate. nAUDC = 1.0 means uniform performance across conditions; lower values indicate condition-dependent failures. \textbf{Threshold:} $\geq$0.85.

\subsubsection{CRG (Compositional Reasoning Gap)}

\begin{equation}
\text{CRG} = 1 - \frac{\sum_i \mathbb{1}[\hat{a}_i = a_i \;\land\; \hat{s}_i = s_i]}{\sum_i \mathbb{1}[\hat{a}_i = a_i]}
\end{equation}
where $\hat{s}_i, s_i$ are predicted and GT safety judgments, computed only on the subset where the model correctly identifies the action ($\hat{a}_i = a_i$).

\textbf{Rationale.} A model that correctly identifies ``worker welding without eye protection'' but reports ``no safety violation'' has a reasoning failure, not a perception failure. CRG isolates this by conditioning on perception success. \textbf{Threshold:} $\leq$0.20.

\textbf{Worked example.} A model correctly identifies B5 (crouching floor work). GT indicates no helmet (unsafe\_act = ``no\_helmet\_in\_work\_zone''). The model predicts unsafe\_act = ``none''. Perception is correct ($\hat{a} = a$), but safety judgment is wrong ($\hat{s} \neq s$) --- this instance increments the CRG numerator.

\subsubsection{Safety Recall and False Alarm Rate}

\begin{align}
\text{Safety Recall} &= \frac{TP}{TP + FN}, \qquad
\text{False Alarm Rate} = \frac{FP}{FP + TN}
\end{align}
where TP = violation correctly detected, FN = violation missed, FP = safe situation falsely flagged, TN = safe situation correctly cleared. Safety judgment is binary (violation present or absent).

\textbf{Rationale.} The two error types have asymmetric deployment consequences: false-safe errors (FN) leave hazards undetected; false-alarm errors (FP) cause operator fatigue. Reporting both exposes which regime each model occupies. \textbf{Threshold:} Recall $\geq$0.90.

\subsubsection{DWA (Distance-Weighted Accuracy)}

\begin{equation}
\text{DWA} = 1 - \frac{1}{N} \sum_{i=1}^{N} d(\hat{a}_i, a_i)
\end{equation}
where $d(\hat{a}_i, a_i)$ is the taxonomic distance:

\begin{table}[htpb]
\centering
\caption{Taxonomic distance function used in DWA. Distances assign smaller 
penalties to semantically related action confusions within the same group and 
larger penalties to errors across distant action groups.}
\label{tab:dwa_distance}
\small
\begin{tabular}{lcc}
\toprule
\textbf{Relationship} & \textbf{Distance} & \textbf{Example} \\
\midrule
Exact match & 0.00 & B1 $\rightarrow$ B1 \\
Same group, different class & 0.33 & B1 $\rightarrow$ B7 \\
Adjacent groups & 0.60 & A1 $\rightarrow$ B1 \\
Distant groups & 0.70--1.00 & A1 $\rightarrow$ D1 \\
\bottomrule
\end{tabular}
\end{table}

\textbf{Rationale.} Standard accuracy treats all misclassifications equally, but confusing ``walking'' (A1) with ``carrying'' (A4) has different safety implications than confusing ``walking'' (A1) with ``crane signaling'' (C1). DWA penalises cross-group confusions more heavily than within-group confusions. \textbf{Threshold:} $\geq$0.80.

\subsubsection{F2-detect (Worker Detection F-beta)}

\begin{align}
TP &= \min(\text{pred}, \text{gt}), \quad FP = \max(0, \text{pred} - \text{gt}), \quad FN = \max(0, \text{gt} - \text{pred}) \\
F_2 &= \frac{5 \cdot P \cdot R}{4P + R}, \quad P = \frac{TP}{TP+FP}, \quad R = \frac{TP}{TP+FN}
\end{align}

$\beta$=2 weights recall 4$\times$ over precision: missing a worker is worse than hallucinating one for safety monitoring. \textbf{Threshold:} $\geq$0.70.

\subsubsection{DRS (Deployment Readiness Score)}

\begin{equation}
\text{DRS} = \frac{1}{5} \sum_{k=1}^{5} \mathbb{1}[\text{metric}_k \text{ passes threshold}_k]
\end{equation}

\textbf{Component thresholds:}

\begin{table}[htpb]
\centering
\caption{Component thresholds used in the DRS. 
Each criterion evaluates a distinct capability required for practical 
industrial deployment, including semantic robustness, environmental stability, 
reasoning reliability, safety sensitivity, and worker detection performance.}
\label{tab:drs_thresholds}
\small
\begin{tabular}{lcl}
\toprule
\textbf{Check} & \textbf{Threshold} & \textbf{Rationale} \\
\midrule
DWA $\geq$ 0.80 & 0.80 & Cross-group confusions must be rare \\
nAUDC $\geq$ 0.85 & 0.85 & Performance must survive real conditions \\
CRG $\leq$ 0.20 & 0.20 & $\leq$20\% reasoning failure on correct perceptions \\
Safety Recall $\geq$ 0.90 & 0.90 & Must catch $\geq$90\% of real violations \\
F2-detect $\geq$ 0.70 & 0.70 & Must reliably detect workers \\
\bottomrule
\end{tabular}
\end{table}

DRS is not a regulatory standard. It is a diagnostic checklist that exposes which capabilities are present and which are missing. Threshold sensitivity ($\pm$10\%) is analyzed in Appendix~\ref{app:drs_sensitivity}.

\subsection{Evaluation Protocol Details}

\subsubsection{Prompt Design}

All 9 evaluation models receive the same zero-shot structured prompt. The system prompts the model as a safety officer reviewing surveillance footage. It specifies the camera distance (7 to 10 meters), the fixed CCTV viewpoint, and the instruction to be detailed and conservative on safety. The user prompt instructs the model to:

\begin{enumerate}[nosep]
\item Identify all visible workers with physical descriptions
\item Classify each worker's primary action from the 25-class taxonomy
\item Provide a free-text description of what the worker is doing
\item Assess spatial context using a fixed tag vocabulary
\item Assess 5 PPE items per worker
\item Classify coordination (solo, coordinated, concurrent)
\item Flag any visible safety violations
\item Report action transitions across the 8 frames
\item Assess occlusion level and source
\item Tag visibility conditions for the scene
\end{enumerate}

The model returns a structured JSON response matching a predefined schema. The full prompt text (approximately 2,500 tokens) is included in the code release.

\subsubsection{Model Configurations}

Table~\ref{tab:app_model_config} lists the evaluated models with provider, API identifier, and access method. All models are evaluated in zero-shot mode with no task-specific fine-tuning. Temperature is set to default (provider-specific). Maximum output tokens are set high enough to accommodate scenes with up to 21 workers.

\begin{table}[h]
\centering
\caption{Model configurations for evaluation. All models were evaluated on the full 1,345-clip set.}
\label{tab:app_model_config}
\scriptsize
\begin{tabular}{llll}
\toprule
\textbf{Model} & \textbf{Provider} & \textbf{API Identifier} & \textbf{Tier} \\
\midrule
GPT-4o & OpenAI & gpt-4o & Frontier \\
GPT-5.4 & OpenAI & gpt-5.4 & Frontier \\
Claude Opus 4.7 & Anthropic & claude-opus-4-7 & Frontier \\
Gemini 2.5 Pro & Google & gemini-2.5-pro & Frontier \\
Gemini 2.5 Flash & Google & gemini-2.5-flash & Frontier \\
Qwen3.5-122B & DeepInfra & Qwen/Qwen3.5-122B-A10B & Large open \\
Llama 4 Maverick & DeepInfra & meta-llama/Llama-4-Maverick-17B-128E & Large open \\
Gemma 4-31B & DeepInfra & google/gemma-4-31B-it & Small/med open \\
Nemotron-12B & DeepInfra & nvidia/Nemotron-Nano-12B-v2-VL & Small/med open \\
\bottomrule
\end{tabular}
\end{table}

\subsubsection{Output Parsing}

VLM outputs are parsed into the evaluation schema using rule-based JSON extraction. The parser first attempts to extract a JSON object from the response. If the response contains markdown code fences, these are stripped before parsing. If JSON parsing fails, the parser attempts to extract key fields using regular expressions. Parse success rate exceeds 98\% across all models. Failed parses are treated as incorrect predictions. The most common parse failure mode is truncated output on clips with many workers, where the response exceeds the model's output token limit and the JSON structure is incomplete.

\subsubsection{Annotation Guidelines Summary}

Annotators received a training session covering the 25-class taxonomy with example clips for each action class. Key instructions included:

\begin{itemize}[nosep]
\item Assign exactly one action code per visible worker based on the primary activity observed across all 8 frames
\item When uncertain between two classes within the same group, prefer the more specific class
\item When uncertain between classes in different groups, flag the clip for expert review
\item Assess PPE items as ``worn'' only if clearly visible; use ``cannot\_determine'' for items below resolution
\item Report safety violations by citing the specific rule code from the facility rule book
\item Layer 2 (up to 5 workers): annotate every visible worker individually
\item Layer 1 (more than 5 workers): annotate scene-level dominant actions and overall compliance
\end{itemize}

Each annotator completed a calibration set of 25 clips with expert feedback before entering the main annotation queue.

\section{Complete Experimental Results}
\label{app:full_results}

This section provides the complete experimental data underlying Section~\ref{sec:results}. The main paper reports aggregate metrics across 9 models; here, we decompose performance along four axes—action class, safety regime, visibility condition, and plant zone—to identify where and why models fail.

\subsection{Per-Class Accuracy}
\label{app:perclass}

The per-class heatmap (Figure~\ref{fig:perclass} in the main paper) shows accuracy for all 25 action classes across 9 models. Here, we provide the detailed per-class analysis summarised in the main paper.

Four patterns emerge from the heatmap:

\begin{itemize}[nosep]
    \item \textbf{Within-group variance exceeds between-group variance.} Group~A (locomotion) contains both the easiest class (A1 walking, 40-78\%) and near-zero classes (A2/A3 climbing). The vertical motion component in climbing is invisible from overhead CCTV --- a fundamental limitation of the camera geometry, not the model.
    \item \textbf{B1 absorbs default predictions.} B1 (operating equipment) accuracy varies widely (28-59\%) because multiple models default to it when uncertain --- the ``B1-magnet'' effect. This inflates both correct and incorrect counts for B1, making its accuracy a noisy indicator of actual recognition ability.
    \item \textbf{Groups C and D are functionally invisible.} Crane operations (C1--C4) and material handling (D1--D3) accuracies varies widely (10-75\%) across all 9 architectures. These are among the most safety-critical activities in the plant --- crane signalling errors and improper lifting cause the most severe injuries --- yet no model recognises them. The failure appears across all evaluated model families and scales, suggesting a data-level or representation gap rather than a prompt engineering problem.
    \item \textbf{Idle-state recognition varies by model bias.} F1 (idle standing) shows 53\% accuracy for Gemma but 7-8\% for Claude and Nemotron. Gemma defaults \emph{to} F1 when uncertain; others default \emph{away} from it. Neither pattern reflects genuine idle-state recognition, and both reflect different default behaviours when the model cannot distinguish the actual activity.
\end{itemize}

\subsection{Safety Operating Modes}
\label{app:safety_modes}

The safety scatter (Figure~\ref{fig:safety_scatter} in the main paper) visualises the two safety operating regimes identified in Section~\ref{sec:model_eval}. Here we provide additional analysis of the clustering pattern.

The scatter plot reveals a structural constraint: no model occupies the upper-left quadrant (high recall, low false alarm rate). Models cluster into two regimes with no intermediate positions. The false-alarm cluster (7 models) achieves 61--88\% recall, but 60--93\% of their safety errors are false alarms, generating unsustainable alert volumes. The false-safe cluster (GPT-4o, Llama~4) produces fewer false alarms, but 63--66\% of their safety errors are missed violations.

The clustering by model family is notable: both OpenAI-lineage models and Llama~4 are false-safe biased, while all other instruction-tuned models are false-alarm biased. This pattern is consistent across action groups and visibility conditions (Appendix~\ref{app:false_safe}), suggesting that these operating modes are model-family specific rather than driven solely by individual visual conditions. The gap between the two clusters --- there are no models with recall between 40\% and 60\% --- indicates that the transition between safety regimes may be discontinuous.

\subsection{DRS Sensitivity Analysis}
\label{app:drs_sensitivity}

Table~\ref{tab:app_drs_sensitivity} tests whether the headline conclusion --- no model reaches DRS~$\geq$~0.60 --- depends on the specific threshold values chosen.

\begin{table}[h]
\centering
\caption{DRS under default and relaxed ($-$10\%) thresholds. The three safety-critical checks (DWA, CRG, safety recall) fail under all threshold settings for all models.}
\label{tab:app_drs_sensitivity}
\small
\begin{tabular}{lcc|cc}
\toprule
\textbf{Model} & \textbf{DRS (default)} & \textbf{DRS ($-$10\%)} & \textbf{New passes} & \textbf{Max possible} \\
\midrule
GPT-4o & 0.40 & 0.40 & --- & 2/5 \\
Gemma 4-31B & 0.40 & 0.40 & --- & 2/5 \\
Llama 4 & 0.40 & 0.40 & --- & 2/5 \\
Qwen3.5-122B & 0.20 & 0.20 & --- & 1/5 \\
Claude Opus & 0.20 & 0.20 & --- & 1/5 \\
GPT-5.4 & 0.20 & 0.20 & --- & 1/5 \\
Gemini Flash & 0.20 & 0.40 & nAUDC & 2/5 \\
Gemini Pro & 0.20 & 0.20 & --- & 1/5 \\
Nemotron & 0.20 & 0.20 & --- & 1/5 \\
\bottomrule
\end{tabular}
\end{table}

Under 10\% relaxation, only one model (Gemini Flash) gains an additional check (nAUDC crosses the relaxed 0.765 threshold). The three checks that drive the deployment gap(DWA, CRG, and safety recall) fail by margins far exceeding 10\%. The best CRG is 0.375 against a 0.20 threshold (87\% above); the best DWA is 0.616 against 0.80 (23\% below); the best safety recall in the false-safe cluster is 0.31 against 0.90 (66\% below). These margins confirm that the conclusion is not an artefact of threshold selection.

\subsection{Degradation by Visibility Condition}
\label{app:degradation}

\begin{figure}[h]
    \centering
    \includegraphics[width=\textwidth]{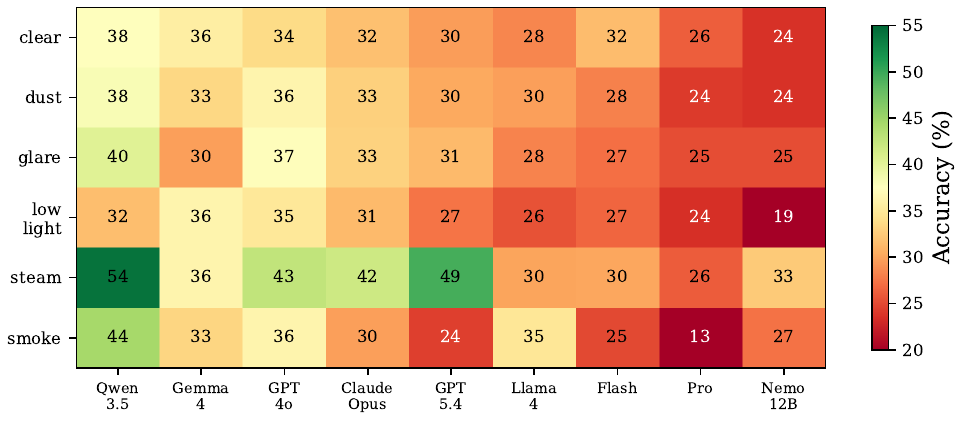}
    \caption{accuracy (\%) by visibility condition across 9 models. Steam and smoke (bottom rows) degrade all models, but the magnitude varies substantially by architecture --- Gemma~4 maintains the most consistent performance while GPT-5.4 shows the sharpest drops.}
    \label{fig:app_degradation}
\end{figure}

Figure~\ref{fig:app_degradation} shows per-model accuracy across 6 natural visibility conditions arising from plant operations.

Degradation patterns are model-specific rather than uniform. Gemma~4 maintains the most stable performance across conditions (nAUDC = 0.912), while GPT-5.4 shows the sharpest drops under steam and smoke (nAUDC = 0.684). This dissociation between accuracy and robustness --- Qwen leads on accuracy (42.6\%), but Gemma leads on stability --- indicates that degradation resilience is a distinct capability not captured by aggregate accuracy.

Within-camera controls confirm that the degradation effect is not a proxy for site difficulty. Cramér's V between camera identity and visibility condition is 0.40—moderate correlation, since some cameras face blast-furnace cooling (steam) or welding zones (smoke). After controlling for camera, the residual degradation effect persists, confirming that visibility conditions independently reduce model performance beyond any site-level confound.

\subsection{Site-Level Accuracy}
\label{app:site_results}

\begin{figure}[h]
    \centering
    \includegraphics[width=\textwidth]{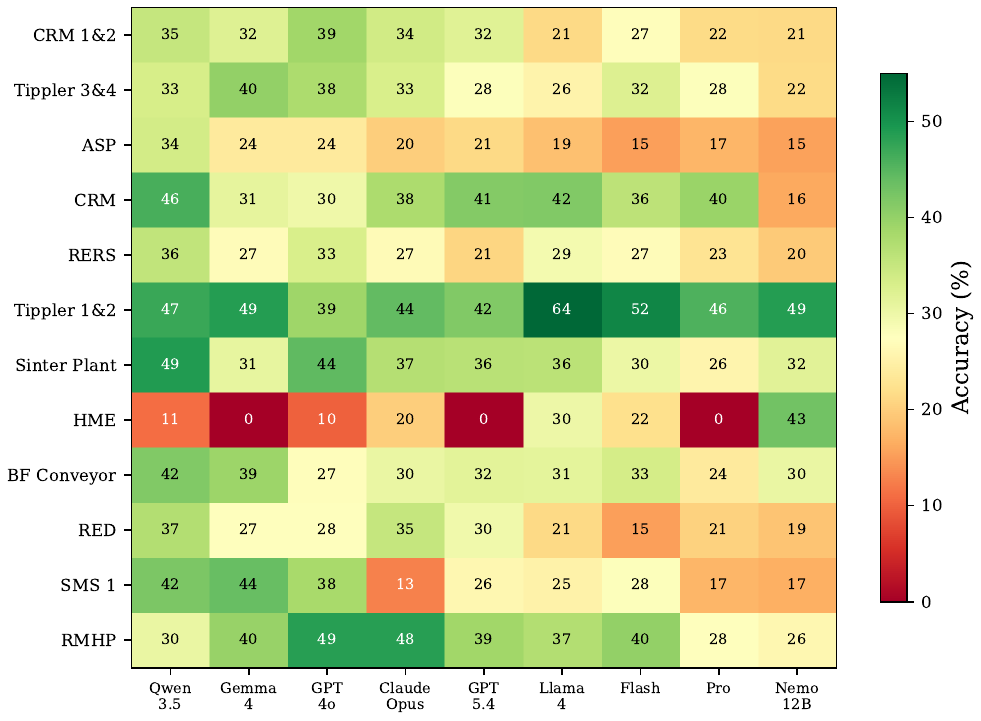}
    \caption{Per-site accuracy (\%) across 12 plant zones and 9 models. Site difficulty is the dominant source of variance (35.9pp spread), exceeding visibility conditions (8.4pp) and worker count (9.4pp).}
    \label{fig:app_site_heatmap}
\end{figure}

Figure~\ref{fig:app_site_heatmap} shows accuracy across 12 plant zones with $\geq$10 evaluation clips.

Site difficulty is the single largest source of performance variance in SteelBench, exceeding visibility conditions (8.4pp spread) and worker count (9.4pp). Accuracy ranges 35.9pp across zones: HME (motor workshop) is among the hardest zones for most models, while Tippler~1\&2 is among the easiest (39--49\%).

Site difficulty has both shared and architecture-specific components. Cross-model Spearman $\rho$ on site rankings averages 0.495, and HME is hard for every model, while Tippler is easy for every model. But substantial architecture-specific variation exists: Claude scores 12.9\% on SMS~1 while Gemma scores 43.8\% on the same zone. This implies that site-level deployment decisions cannot be made based on aggregate benchmark scores, and a model that performs adequately in one zone may fail in another, even within the same facility.

\subsection{Calibration Detail}

\begin{table}[h]
\centering
\caption{Per-model calibration. Eight of 9 models report 75--90\% confidence regardless of actual accuracy (28--43\%), producing confidence-accuracy gaps of 40--61pp. Claude Opus is the sole exception.}
\label{tab:app_calibration}
\small
\begin{tabular}{lcccc}
\toprule
\textbf{Model} & \textbf{Accuracy} & \textbf{Avg Conf.} & \textbf{Conf--Acc Gap} & \textbf{ECE} \\
\midrule
Qwen3.5-122B & 42.6\% & 90.1\% & 47.5pp & 0.475 \\
Gemma 4-31B & 39.2\% & 84.5\% & 45.3pp & 0.453 \\
GPT-4o & 38.8\% & 85.7\% & 46.9pp & 0.469 \\
Claude Opus 4.7 & 37.4\% & 66.0\% & 28.6pp & \textbf{0.286} \\
GPT-5.4 & 35.5\% & 75.4\% & 39.9pp & 0.399 \\
Llama 4 & 34.5\% & 84.8\% & 50.2pp & 0.502 \\
Gemini Flash & 33.6\% & 89.8\% & 56.2pp & 0.562 \\
Gemini Pro & 29.5\% & 88.5\% & 59.0pp & 0.590 \\
Nemotron-12B & 27.8\% & 88.3\% & 60.5pp & 0.605 \\
\bottomrule
\end{tabular}
\end{table}

The calibration table reveals a near-universal pattern: eight of 9 models report 75--90\% confidence regardless of their actual accuracy, which ranges from 27.8\% to 42.6\%. The resulting confidence-accuracy gaps (40--61pp) mean these models express high certainty about predictions that are wrong more often than right.

Claude Opus~4.7 is the sole exception, reporting 66.0\% mean confidence --- the lowest in the pool --- and achieving the best ECE (0.286) as a result. Notably, Claude's lower confidence does not reflect lower capability: its accuracy (37.4\%) is within 5pp of the top model. The gap between Claude and the rest suggests that calibration quality may be a model-family property rather than a byproduct of model capability --- a finding with implications for human-in-the-loop systems, where well-calibrated uncertainty is as valuable as accuracy.

\subsection{Human Reference Methodology}                                                                                                                                                                  
  \label{app:human_baseline}                                       
                                                                                                                                                                                                            
  The 84.6\% human reference accuracy and the 42.0pp human-model gap are                                                                                                                                    
  central claims. This section documents the methodology in full.                                                                                                                                           
                                                                                                                                                                                                            
  \subsubsection{Who Are the Humans}                                                                                                                                                                        
                                                                                                                                                                                                            
  Five Tier-1 annotators and two domain experts participated. Tier-1                                                                                                                                        
  annotators received a training session covering the 25-class action                                                                                                                                       
  taxonomy with example clips for each class, followed by a 25-clip                                                                                                                                         
  calibration exercise with expert feedback. Domain experts have                                                                                                                                            
  backgrounds in industrial safety and access to the facility's safety                
  rule documentation. Table~\ref{tab:human_ref_annotators} summarises                                                                                                                                       
  per-annotator contribution.                                                         
                                                                   
  \begin{table}[h]                                                                                                                                                                                          
  \centering                                                                                                                                                                                                
  \caption{Per-annotator contribution to the human reference evaluation.                                                                                                                                    
  Proper-chain clips are those where the expert reviewed Tier-1 work (not                                                                                                                                   
  VLM output directly).}                                                                                                                                                                                    
  \label{tab:human_ref_annotators}                                                                                                                                                                          
  \small                                                                                                                                                                                                    
  \begin{tabular}{lccc}                                                               
  \toprule                                                                                                                                                                                                  
  \textbf{Annotator} & \textbf{Total clips} & \textbf{In proper-chain} & \textbf{Action accuracy vs.\ expert} \\                                                                                            
  \midrule                                                                                                                                                                                                  
  Annotator 1 & 434 & 83 & 86.7\% \\                                                                                                                                                                        
  Annotator 2 & 443 & 45 & 82.2\% \\                                                                                                                                                                        
  Annotator 3 & 386 & 39 & 84.6\% \\                                                                                                                                                                        
  Annotator 4 & 303 & 38 & 81.6\% \\                                                                                                                                                                        
  \midrule                                                                                                                                                                                                  
  All (proper-chain) & --- & 174 clips / 370 pairs & 84.6\% \\                        
  \bottomrule                                                                                                                                                                                               
  \end{tabular}                                    
  \end{table}                                                                                                                                                                                               
                                                                                                                                                                                                            
  \subsubsection{What Is the Expert Reference}                     
                                                                                                                                                                                                            
  The evaluation uses a \emph{proper-chain} design. The annotation                    
  pipeline proceeds as VLM pre-fill $\rightarrow$ Tier-1 correction                                                                                                                                         
  $\rightarrow$ expert verification. In the proper-chain subset, experts
  reviewed Tier-1 annotations and either accepted or corrected each field.                                                                                                                                  
  Expert judgment is treated as the reference standard.                                                                                                                                                     
                                                                                                                                                                                                            
  This is distinct from the \emph{VLM-sourced} subset (262 clips), where                                                                                                                                     
  experts edited VLM output directly without a Tier-1 intermediate. The               
  human baseline is computed only on proper-chain clips to avoid                                                                                                                                            
  double-counting VLM influence.                                                      
                                                                                                                                                                                                            
  \subsubsection{Were Humans Blind to Model Outputs?}                                 
                                                                                                                                                                                                            
  Tier-1 annotators saw VLM pre-fill suggestions as a starting point and
  corrected them. They were not blind to VLM outputs. This reflects the                                                                                                                                     
  realistic annotation workflow: the 84.6\% measures human performance                
  under the actual conditions used to construct the benchmark, not under                                                                                                                                    
  stripped-down testing conditions.      
  Separate blind annotations (102 clips where Tier-1 annotators received                                                                                                                                     
  no VLM pre-fill) exist for measuring anchoring bias. Blind-condition                                                                                                                                      
  accuracy against expert reference is lower, consistent with the task                
  being genuinely difficult at surveillance distance without any starting                                                                                                                                   
  point.       
  \subsubsection{How Does 84.6\% Relate to $\kappa$=0.82}                                      
  Both are computed on the same 370 person-level pairs from 174                                                                                                                                             
  proper-chain clips. The 84.6\% is the observed agreement rate ($p_o$).              
  Cohen's $\kappa$=
  0.82 adjusts for chance agreement across 25 action                                                                                                                                       
  classes. The high $\kappa$ confirms that agreement is not driven by                                                                                                                                       
  class imbalance (if both annotators defaulted to the majority class,                                                                                                                                      
  $p_o$ would be high but $\kappa$ would be low).                                              
  \subsubsection{How Was Disagreement Resolved}                                              
  When Tier-1 and the expert disagreed on an action code, the expert's judgment                                                                                                                                   
  was retained as ground truth. Experts had access to facility safety                                                                                                                                       
  documentation and could review all 8 frames at full resolution. For the                                                                                                                                   
  15.4\% of pairs where Tier-1 and expert disagreed, the most common                                                                                                                                        
  pattern was within-group confusion (e.g., B1 vs.\ B7) rather than                   
  cross-group error.                                                                                                                                                                                        
                                                                                      
  \subsubsection{Unit of Measurement}                                                       
  The 84.6\% is \textbf{per-person action accuracy} on the 25-class                                                                                                                                         
  taxonomy. Each visible worker is one comparison unit. A clip with 3                                                                                                                                       
  workers contributes 3 pairs. Layer-1 clips (scene-level annotation)                                                                                                  contribute 1 pair using the dominant action code. The 370 pairs come                                                                                                                                      
  from 174 clips because many clips contain multiple workers.                                                                                                                                               
                                                                              
\section{Extended Ablation Data}
\label{app:additional_ablations}

This section provides the full metrics and additional analyses supporting
the ablation study in Section~\ref{sec:ablations}. The main paper reports
headline findings. Here we provide the complete tables, per-model curves,
and supplementary decompositions not included in the main text due to space.

\subsection{Protocol validations.} All 9 models score higher on anchored GT than
on blind GT, with a mean gain of 6.6pp. Relative robustness rankings remain
stable across provenance slices. Worker count causes a gradual degradation of
about 5pp per additional worker rather than a sharp cliff. Full details are
in Appendices~\ref{app:blind_anchored}, \ref{app:drs_sensitivity},
and~\ref{app:worker_scaling}.

\subsection{Prompt Sensitivity: Full Metrics}
\label{app:prompt_sensitivity}

Table~\ref{tab:app_prompt_full} reports all evaluation metrics across three prompt variants (V1: base protocol with taxonomy and safety rules; V2: open-description with no taxonomy; V3: structured observation with taxonomy retained). V1 results use the 150-clip subset of the main evaluation data; V2 and V3 were run as separate inference runs.

\begin{table}[h]
\centering
\caption{Full prompt sensitivity results (150 stratified clips). Accuracy is stable between taxonomy-preserving variants V1 and V3 ($\leq$2.5pp). V2 accuracy is lower due to mapper noise from open-form descriptions. B1 prediction frequency drops 2--4$\times$ under V2, revealing taxonomy force-fitting.}
\label{tab:app_prompt_full}
\small
\begin{tabular}{llccccccc}
\toprule
\textbf{Model} & \textbf{Variant} & \textbf{Acc} & \textbf{CRG} & \textbf{B1\%} & \textbf{F1\%} & \textbf{ECE} & \textbf{Conf} & \textbf{Tokens} \\
\midrule
Gemma 4 & V1 (base) & 29.5 & 0.580 & 14.1 & 32.5 & 0.548 & 84.3 & --- \\
Gemma 4 & V2 (open) & 21.7 & 0.346 & 6.2 & 40.8 & 0.854 & 85.4 & 1,067 \\
Gemma 4 & V3 (structured) & 32.0 & 0.500 & 17.7 & 34.6 & 0.500 & 82.0 & 1,377 \\
\midrule
GPT-4o & V1 (base) & 33.7 & 0.451 & 25.5 & 18.5 & 0.519 & 85.6 & --- \\
GPT-4o & V2 (open) & 24.0 & 0.390 & 6.4 & 18.7 & 0.872 & 87.2 & 871 \\
GPT-4o & V3 (structured) & 34.8 & 0.404 & 20.7 & 12.8 & 0.502 & 84.9 & 1,070 \\
\bottomrule
\end{tabular}
\end{table}

Three findings emerge beyond the headline $\pm$2.5pp accuracy stability. First, \textbf{CRG improves under V3} for both models ($-$0.05 to $-$0.08), suggesting that structured observation partially addresses reasoning failures. CRG nevertheless remains well above the 0.20 deployment threshold, indicating that the reasoning gap is not prompt-dependent.

Second, \textbf{V2 exposes taxonomy force-fitting.} B1 prediction frequency drops from 14--25\% to 6\% when the taxonomy is removed. Without the B1 label available, models describe the same scenes as ``standing watching'' or ``idle,'' which maps to F1 or B7 rather than B1. This confirms that the B1-magnet bias (Appendix~\ref{app:perclass}) is partly an artefact of the closed-vocabulary prompt design: models select B1 as a catch-all when no better option fits, not because they recognise equipment operation.

Third, \textbf{V2 shifts safety behaviour.} Gemma flips from false-alarm biased (V1) to false-safe biased (V2), revealing that the injected safety rules in V1/V3 drive over-flagging behavior. The model's intrinsic safety reasoning, without explicit rule injection, underdetects violations. This has implications for prompt design in deployment: safety rule injection improves recall at the cost of false alarms, and the optimal balance depends on operational tolerance.

V2 free-form action descriptions were mapped to the 25-class taxonomy using GPT-4o-mini (1,227 descriptions, 100\% parse success). V2 accuracy includes mapper noise and should not be directly compared to V1/V3 without accounting for this additional processing step.

\subsection{Frame Density Analysis}
\label{app:frame_density}

\begin{figure}[h]
    \centering
    \includegraphics[width=0.75\textwidth]{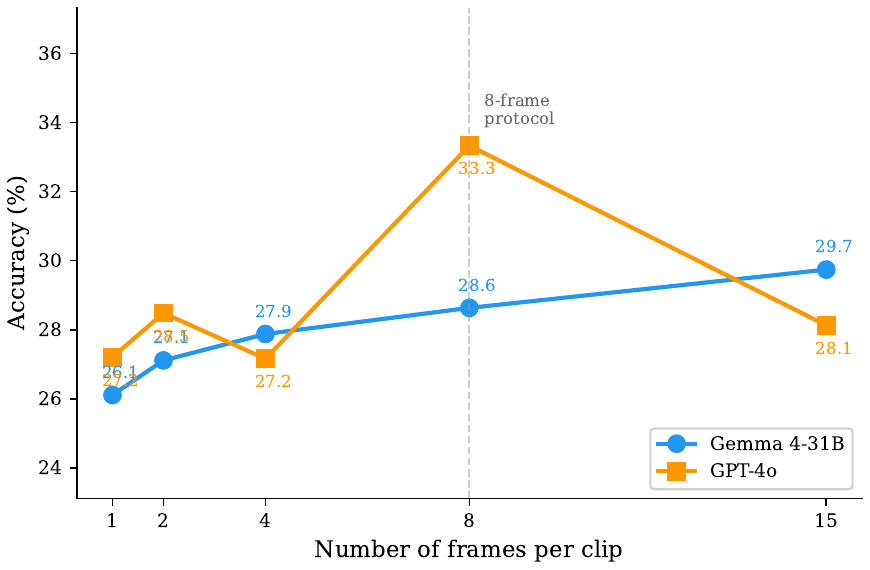}
    \caption{Accuracy by frame count (150 stratified clips). Gemma~4 improves gradually from 1 to 15 frames (+3.6pp total). GPT-4o peaks at 8 frames and declines at 15, suggesting context-length or attention degradation with many images. The dashed line marks the 8-frame protocol used in the main evaluation.}
    \label{fig:app_frame_density}
\end{figure}

Figure~\ref{fig:app_frame_density} shows action accuracy as a function of frames per clip for two representative models.

The two models exhibit qualitatively different scaling behaviours. \textbf{Gemma~4} improves monotonically: 26.1\% (1f) $\rightarrow$ 27.1\% (2f) $\rightarrow$ 27.9\% (4f) $\rightarrow$ 28.6\% (8f) $\rightarrow$ 29.7\% (15f). The gain is gradual and diminishing, and each doubling of frames adds roughly 1pp. Even at 15 frames (1\, fps for the full 15-second clip), the total improvement over a single centre frame is only 3.6pp, confirming that the bottleneck is fine-grained action discrimination at surveillance distance, not temporal sampling density.

\textbf{GPT-4o} peaks at 8 frames (33.3\%) and \emph{declines} at 15 frames (28.1\%, $-$5.2pp). This non-monotonic pattern suggests that GPT-4o's visual processing degrades with increasing input image count, possibly due to context-length effects or attention dilution across 15 high-resolution frames. The 8-frame protocol is therefore not only cost-effective but architecture-appropriate: it provides the best performance for the strongest frontier model while maintaining competitive performance for open-weight models.

\subsection{CRG Decomposition by Action Group}
\label{app:crg_decomposition}

Table~\ref{tab:app_crg_group} decomposes CRG by action group, revealing that safety reasoning failures are not uniform across activity types. Each cell reports CRG and the number of perception-correct instances ($n$) used to compute it.

\begin{table}[h]
\centering
\caption{CRG by action group, reported as CRG ($n$). Higher CRG = worse safety reasoning. Groups~C/D have very small $n$ because perception accuracy on these classes is near-zero (130 and 147 GT instances exist respectively --- the small $n$ reflects model failure to recognise the action, not data sparsity).}
\label{tab:app_crg_group}
\scriptsize
\begin{tabular}{lcccccc}
\toprule
\textbf{Model} & \textbf{A} & \textbf{B} & \textbf{C} & \textbf{D} & \textbf{E} & \textbf{F} \\
\midrule
Qwen 3.5 & .53 (213) & .58 (319) & .50 (4) & .00 (2) & .69 (36) & .62 (159) \\
Gemma 4 & .52 (178) & .56 (298) & .00 (1) & --- (0) & .33 (3) & .54 (167) \\
GPT-4o & .40 (194) & .41 (347) & .25 (4) & .00 (5) & .32 (19) & .39 (130) \\
Claude Opus & .52 (198) & .62 (369) & .20 (5) & .00 (1) & .67 (21) & .58 (43) \\
GPT-5.4 & .46 (235) & .55 (280) & .75 (4) & .00 (2) & .36 (14) & .48 (89) \\
Llama 4 & .33 (196) & .38 (312) & .50 (4) & .00 (1) & .38 (13) & .32 (71) \\
Flash & .44 (153) & .52 (276) & .25 (4) & .00 (2) & .57 (28) & .51 (96) \\
Pro & .49 (145) & .52 (259) & .40 (5) & --- (0) & .43 (44) & .43 (47) \\
Nemotron & .38 (93) & .60 (281) & .50 (4) & .00 (6) & .68 (19) & .36 (39) \\
\bottomrule
\end{tabular}
\end{table}

Two patterns are robust despite the small-$n$ limitations in Groups~C/D. First, \textbf{Groups~E (communication) and B (stationary work) consistently show the highest CRG} across most models (0.38--0.69 for E, 0.38--0.62 for B). These are activities with context-dependent safety implications: a worker communicating near moving equipment may or may not be in danger, depending on the spatial context and PPE state. In contrast, Group~D actions (material handling) show CRG = 0.00 wherever $n > 0$ --- when a model correctly identifies material handling, the safety implication is unambiguous.

Second, \textbf{GPT-4o and Llama~4 consistently show the lowest CRG} across groups with sufficient $n$, while Qwen and Claude show the highest. This reinforces the accuracy-CRG pattern from Section~\ref{sec:model_eval}: models that perceive more actions correctly tend to reason about their safety implications less reliably.

\subsection{False-Safe Asymmetry}
\label{app:false_safe}

Table~\ref{tab:app_false_safe} reports the direction of safety errors per model, extending the safety operating mode analysis from Section~\ref{sec:model_eval} with exact counts.

\begin{table}[h]
\centering
\caption{Safety error direction per model. GPT-4o and Llama~4 miss 63--66\% of real violations when they make safety errors. All other models over-flag safe situations.}
\label{tab:app_false_safe}
\small
\begin{tabular}{lrrrl}
\toprule
\textbf{Model} & \textbf{False-Safe} & \textbf{False-Alarm} & \textbf{Total Errors} & \textbf{Bias} \\
\midrule
Qwen 3.5 & 82 (7\%) & 1,039 (93\%) & 1,121 & False-alarm \\
Gemma 4 & 161 (16\%) & 854 (84\%) & 1,015 & False-alarm \\
GPT-4o & 510 (66\%) & 260 (34\%) & 770 & \textbf{False-safe} \\
Claude Opus & 113 (10\%) & 1,001 (90\%) & 1,114 & False-alarm \\
GPT-5.4 & 238 (23\%) & 810 (77\%) & 1,048 & False-alarm \\
Llama 4 & 507 (63\%) & 304 (37\%) & 811 & \textbf{False-safe} \\
Flash & 252 (25\%) & 746 (75\%) & 998 & False-alarm \\
Pro & 176 (17\%) & 880 (83\%) & 1,056 & False-alarm \\
Nemotron & 186 (20\%) & 727 (80\%) & 913 & False-alarm \\
\bottomrule
\end{tabular}
\end{table}

The deployment implications of each regime differ fundamentally. A false-safe model deployed for safety monitoring would miss the majority of real violations—unsafe acts by workers would go undetected. A false-alarm model would generate overwhelming alert volumes but would rarely miss a real violation. Neither regime is acceptable for autonomous deployment, but the false-alarm mode is operationally safer because real violations are still caught, even at the cost of alert fatigue. The 7:2 split between regimes and the absence of any model in the balanced middle confirms the structural constraint identified in the main paper (Figure~\ref{fig:safety_scatter}).

\subsection{Worker-Count Scaling}
\label{app:worker_scaling}

\begin{figure}[h]
    \centering
    \includegraphics[width=0.75\textwidth]{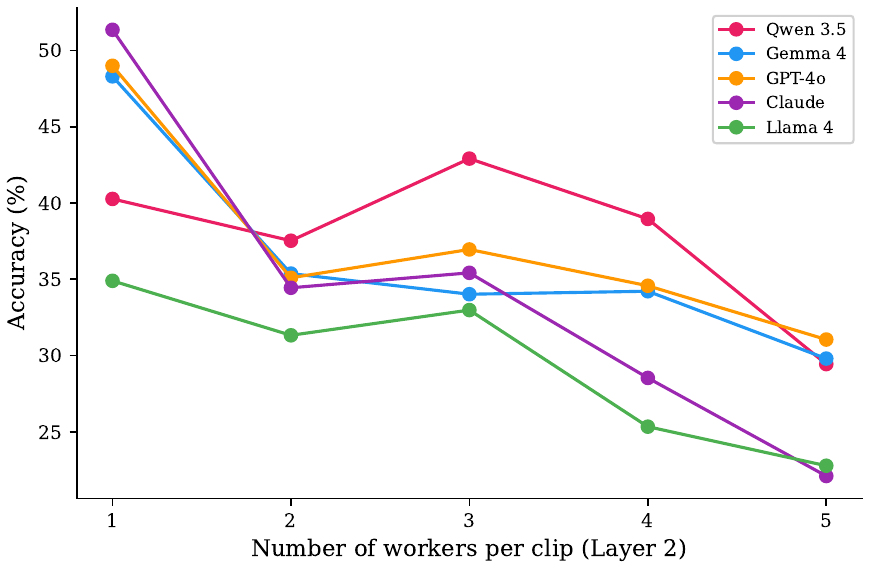}
    \caption{Layer~2 accuracy by worker count (1--5) for top 5 models. The sharpest drop is from 1 to 2 workers (${\sim}$10pp); subsequent workers add ${\sim}$3pp degradation each. No model shows a cliff—the decline is gradual.}
    \label{fig:app_worker_scaling}
\end{figure}

Figure~\ref{fig:app_worker_scaling} shows accuracy as a function of worker count within Layer~2 (per-person annotation, 1--5 workers).

Accuracy decays approximately 5pp per additional worker across all models, but the decay is not uniform. The sharpest drop occurs from 1 to 2 workers (${\sim}$10pp): adding a second person immediately creates ambiguity about which worker is performing which action. Beyond 2 workers, the decline is gentler (${\sim}$ 3 pp per additional worker). At 5 workers, all models fall below 30\% accuracy, approaching a random baseline for the 25-class taxonomy.

The gradual, smooth decline has two implications. First, it confirms that SteelBench tests action discrimination difficulty rather than a counting problem --- if worker count were the bottleneck, we would expect a sharp cliff rather than a smooth decline. Second, it validates the Layer~1/2 boundary at 5 workers (Appendix~\ref{app:pipeline}): beyond this point, per-person accuracy is low enough that scene-level annotation provides a more reliable evaluation signal.

\subsection{GT Provenance: Blind vs Anchored Accuracy}
\label{app:blind_anchored}

\begin{figure}[h]
    \centering
    \includegraphics[width=\textwidth]{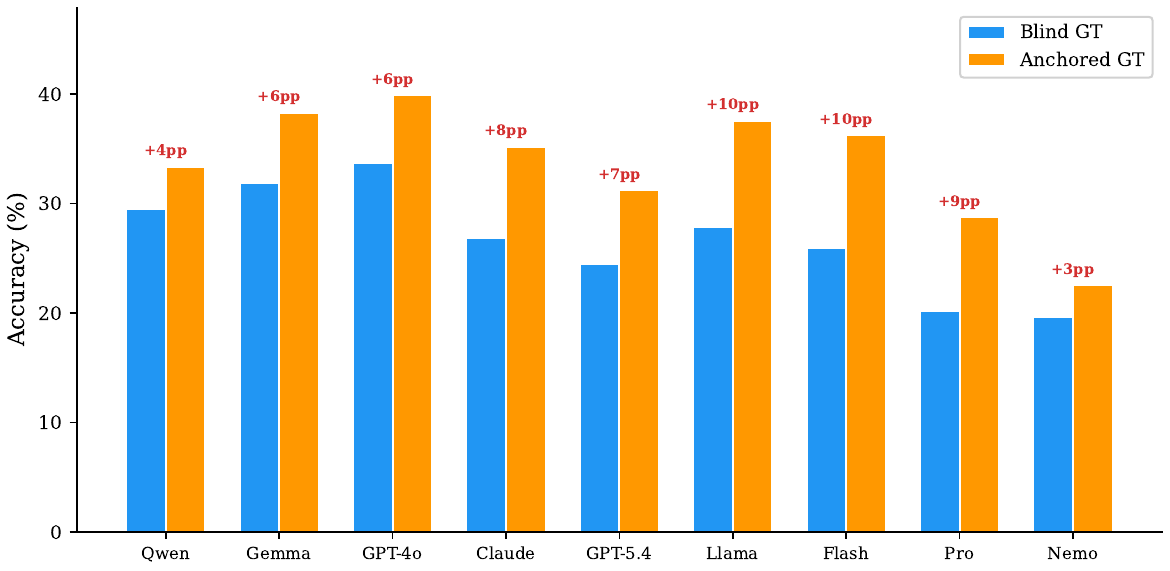}
    \caption{Action accuracy on blind vs anchored GT across all 9 models. Every model scores higher on anchored GT, with a mean inflation of +6.6pp. The universal direction confirms that GT provenance distortion is not family-specific.}
    \label{fig:app_blind_anchored}
\end{figure}

Figure~\ref{fig:app_blind_anchored} shows that all 9 models score higher on anchored GT (where annotators saw VLM pre-fills) than on blind GT (where they did not).

The mean inflation is +6.6pp, ranging from +2pp (Nemotron) to +10pp (Gemini Flash). Two aspects of this result are notable. First, the inflation is not restricted to the Qwen family that shares lineage with the annotation VLM --- every evaluated model benefits. This indicates that the distortion is a property of model-influenced GT evaluated by VLMs with overlapping training distributions, not a family-specific contamination effect.

Second, the magnitude of inflation varies across models in a pattern that does not correlate with accuracy. Gemini Flash shows the largest gap (+10pp) despite mid-range accuracy (33.6\%), while Nemotron shows the smallest gap (+2pp) with the lowest accuracy (27.8\%). This variation suggests that some model families are more sensitive to GT provenance effects than others --- a confound that would be invisible without provenance-stratified evaluation. Blind GT provides a more conservative evaluation surface.

\subsection{Bootstrap Confidence Intervals}                                                                                                                                 \label{app:bootstrap} 

Table~\ref{tab:bootstrap_ci} reports 95\% bootstrap confidence intervals on weighted accuracy (10,000 clip-level resamples with replacement). All intervals span $\pm$2pp or less. Models separated by more than 4pp have non-overlapping intervals, confirming that ranking differences are statistically meaningful. The 42.0pp human-model gap is 20$\times$ wider than any single interval. 

For per-condition nAUDC, the four largest visibility bins contain $\geq$217 clips: clear (636), dust (608), glare (473), and low\_light (217). The rarest conditions, steam ($n$=50) and smoke ($n$=19), have limited support and should be interpreted cautiously.                                                                                                                                              
                                                                                      
  \begin{table}[h]                                 
  \centering                                                                                                                                                                                                
  \caption{Bootstrap confidence intervals on weighted action accuracy
  (10,000 clip-level resamples). Human reference accuracy is 84.6\%.}                                                                                                                                       
  \label{tab:bootstrap_ci}                                                                                                                                                                                  
  \small                                                                                                                                                                                                    
  \begin{tabular}{lccc}                                                                                                                                                                                     
  \toprule                                                                            
  \textbf{Model} & \textbf{Accuracy (\%)} & \textbf{95\% CI} & \textbf{$\pm$ (pp)} \\                                                                                                                       
  \midrule                                                                            
  Qwen 3.5-122B    & 42.6 & [40.5, 44.8] & 2.1 \\                                                                                                                                                           
  Gemma 4-31B      & 39.2 & [37.1, 41.4] & 2.1 \\                                                                                                                                                           
  GPT-4o           & 38.8 & [36.7, 40.8] & 2.1 \\                                                                                                                                                           
  Claude Opus 4.7  & 37.4 & [35.4, 39.5] & 2.0 \\                                                                                                                                                           
  GPT-5.4          & 35.5 & [33.5, 37.6] & 2.0 \\                                                                                                                                                           
  Llama 4 Maverick & 34.5 & [32.6, 36.6] & 2.0 \\        
  Gemini 2.5 Flash & 33.6 & [31.6, 35.7] & 2.1 \\                                                                                                                                                           
  Gemini 2.5 Pro   & 29.5 & [27.6, 31.5] & 1.9 \\                                     
  Nemotron-12B     & 27.8 & [25.9, 29.8] & 1.9 \\                                                                                                                                                           
  \bottomrule                                            
  \end{tabular}                                                                                                                                                                                             
  \end{table}

\subsection{Cross-Domain Evaluation}                                                                                                                                                                      
  \label{app:domain_diversity}                                                                                                                                                                              
                                                                                                                                                                                                            
  An integrated steel plant contains functionally distinct production areas with different equipment, lighting, camera positions, and worker activities. We group 16 zones into 6 industrial domains and evaluate each model separately on each domain                                
  (Table~\ref{tab:domain_accuracy}).                                                                                                                                                                        
                                                                                                                                                                                                            
  \begin{table}[h]                                                                                                                                                                                          
  \centering                                                                                                                                                                                                
  \caption{Weighted action accuracy (\%) by industrial domain. Each domain has distinct camera geometry, visibility conditions, and                                                                                                                                         
  action-class mix (16--25 of 25 classes per domain). Spread is the domain difference between the highest and lowest domain accuracy for each
  model.}                                                                                                                                                                                                   
  \label{tab:domain_accuracy}                                                                                                                                                                               
  \scriptsize                                                      
  \begin{tabular}{lrcccccccccc}                                                                                                                                                                             
  \toprule                                                                                                                                                                                                  
  \textbf{Domain} & \textbf{$n$} & \rotatebox{70}{Qwen 3.5} & \rotatebox{70}{Gemma 4} & \rotatebox{70}{GPT-4o} & \rotatebox{70}{Claude Opus} & \rotatebox{70}{GPT-5.4} & \rotatebox{70}{Llama 4} &
  \rotatebox{70}{Gem.\ Flash} & \rotatebox{70}{Gem.\ Pro} & \rotatebox{70}{Nemotron} \\                                                                                                                     
  \midrule                                                                                                                                                                                                  
  Rolling Mill   & 634 & 43.2 & 41.0 & 41.8 & 43.3 & 40.5 & 35.0 & 35.3 & 32.8 & 27.6 \\                                                                                                                    
  Raw Material   & 571 & 41.9 & 42.5 & 41.2 & 40.6 & 35.1 & 35.5 & 37.8 & 31.8 & 30.0 \\                                                                                                                    
  Ironmaking     & 454 & 44.7 & 36.8 & 37.7 & 36.8 & 35.1 & 34.6 & 31.2 & 26.9 & 33.3 \\                                                                                                                  
  Steel Making   & 345 & 42.0 & 38.6 & 33.9 & 27.6 & 32.1 & 31.1 & 28.0 & 23.7 & 22.9 \\                                                                                                                    
  Utilities      & 421 & 41.3 & 35.0 & 36.4 & 33.7 & 31.2 & 34.8 & 32.3 & 29.2 & 24.5 \\
  Chemical       &  33 & 36.4 & 39.4 & 36.4 & 30.3 & 45.5 & 42.4 & 40.6 & 28.1 & 12.1 \\                                                                                                                    
  \midrule                                                                                                                                                                                                
  \textbf{Spread (pp)} & & 8.4 & 7.5 & 7.9 & 15.7 & 14.2 & 11.4 & 12.6 & 9.1 & 21.2 \\                                                                                                                      
  \bottomrule                                                                                                                                                                                             
  \end{tabular}                                                                                                                                                                                             
  \end{table}  

  Three observations support internal domain diversity as a proxy for                                                                                                                                       
  cross-facility variation. First, the six domains cover different camera
  distances (overhead cranes vs.\ ground-level conveyor views), lighting                                                                                                                                    
  (open-air tipplers vs.\ enclosed rolling mill bays), and degradation                                                                                                                                      
  sources (blast-furnace dust and smoke vs.\ rolling-mill steam and                                                                                                                                         
  glare). Second, each domain draws from a different subset of the 25                                                                                                                       
  action classes, with coverage ranging from 16 classes (Chemical) to all                                                                                                                                   
  25 (Steel Making). Third, cross-domain accuracy spread reaches 8--21pp                                                                                                                                    
  per model, comparable to the cross-dataset spread reported in transfer                                                                                                                                    
  studies on internet video benchmarks. This variation confirms that zone                                                                                                                                   
  identity produces a meaningful evaluation signal and that the benchmark                                                                                                                                     
  tests more than a single camera configuration.                                                                                                                                                          
                                                                                                                                                                                                            
  We acknowledge that cross-facility generalisation remains untested.                                                                                                                                     
  A multi-plant extension would strengthen external validity. The 16-zone,                                                                                                                                  
  6-domain structure within a single integrated facility is the strongest             
  diversity argument available from a single-plant dataset.

\subsection{Sim-to-Real Performance Gap}
  \label{app:sim_real}

  IndustryEQA~\citep{industryeqa} evaluates VLMs on simulated warehouse
  footage generated in NVIDIA Isaac Sim with scripted OSHA-inspired
  hazards. SteelBench evaluates real operational CCTV with naturally
  occurring activities and violations. Table~\ref{tab:sim_real} compares
  overlapping models across both benchmarks.

  \begin{table}[h]
  \centering
  \caption{Performance comparison on simulated (IndustryEQA) vs.\ real
  (SteelBench) industrial footage. IndustryEQA scores are directly accurate
  from the original paper. The consistent drop indicates that the simulated evaluation overestimates real-world capability.}
  \label{tab:sim_real}
  \small
  \begin{tabular}{lccc}
  \toprule
  \textbf{Model} & \textbf{IndustryEQA (\%)} & \textbf{SteelBench (\%)} & \textbf{Gap (pp)} \\
  \midrule
  GPT-4o           & 57.2 & 38.8 & $-$18.4 \\
  Gemini 2.5 Flash & 55.8 & 33.6 & $-$22.2 \\
  \bottomrule
  \end{tabular}
  \end{table}

  The gap reflects several sources of difficulty absent in the simulation.
  Real CCTV footage contains environmental degradation (dust, steam,
  smoke, glare) that cannot be fully replicated synthetically. Camera
  distance (7--10 meters overhead) reduces workers to 80--200 pixel
  figures, far below the resolution of rendered avatars. Activity
  distributions follow genuine operational patterns rather than scripted
  scenarios, producing a natural long-tail distribution with rare but
  safety-critical classes. These differences confirm that benchmarks built
  on real operational footage test capabilities that simulated environments
  do not exercise.

\section{Datasheet for Datasets}
\label{app:datasheet_full}

This datasheet follows the framework proposed by \citet{gebru2021datasheets}.

\subsection{Motivation}

\paragraph{Why was the dataset created?}
SteelBench was created to provide a diagnostic benchmark for evaluating vision-language models on real industrial surveillance footage. Existing VLM benchmarks use curated internet videos or synthetic scenes that do not capture the environmental degradation, distant viewpoints, multi-worker coordination, and safety-rule reasoning required in operational settings. SteelBench fills this gap with footage from an active steel plant.

\paragraph{What tasks was it designed for?}

The primary task is structured activity recognition under industrial conditions: classifying worker actions from a 25-class taxonomy, assessing PPE compliance, evaluating spatial context, and detecting safety violations. A secondary contribution is the provenance-aware annotation audit, which measures how VLM-assisted labelling affects evaluation outcomes.

\paragraph{Who created the dataset?}
The paper authors created the dataset. 

\paragraph{Who funded its creation?}

Institutional funding and non-monetary support in the form of annotations and safety rules, verification, and random auditing of 200 annotation inputs from the Safety Engineering Department, Bokaro Steel Plant, Steel Authority of India Limited(A Govt of India Enterprise) and no external grants.

\paragraph{Has it been used for tasks already?}
This paper presents the first analysis. Results for 9 VLMs are reported in Section~\ref{sec:results}.

\subsection{Composition}

\paragraph{What are the instances?}
Each instance is a up to 15-second video clip from fixed CCTV cameras, represented by 8 frames sampled at fixed temporal positions (0\%, 14\%, 28\%, 43\%, 57\%, 71\%, 86\%, 100\% of clip duration). Frames are at 1080p resolution. Each clip contains one or more visible workers performing industrial activities.

\paragraph{How many instances are there?}
1,345 evaluation clips drawn from 10,000+ candidate clips extracted from 149 hours of source videos recorded by 64 cameras across 16 plant zones. The clips contain 2,208 person-level annotation instances.

\paragraph{What data does each instance consist of?}
Each clip includes: (1) 8 JPEG frames with Gaussian face blur applied; (2) structured annotations covering action classification, PPE assessment (5 items per worker), spatial context (12 tags), scene type (SA/MAI/MAC), visibility conditions (6 categories, multi-label), safety violations with rule codes, occlusion level, and action transitions; (3) annotation provenance metadata tracking which fields were modified from VLM pre-fill.

\paragraph{Are there labels?}

Yes. Each worker is labelled with a primary action code from 25 classes in 6 groups (A: Locomotion, B: Stationary Work, C: Crane and Equipment, D: Material Handling, E: Social, F: Idle). PPE compliance is labeled per item as \texttt{worn}, \texttt{not\_worn}, \texttt{cannot\_determine}, or \texttt{not\_applicable}. Safety violations are referenced by specific facility(Bokaro Steel Plant) rule codes.

\paragraph{Are there multiple types of instances?}
Yes. Clips with $\leq$5 workers receive per-person annotation (Layer 2, 805 clips). Clips with $>$5 workers receive scene-level annotation (Layer 1, 540 clips).

\paragraph{Are relationships between instances explicit?}
No direct relationships. Clips from the same camera or zone can be identified through the clip ID.

\paragraph{Is the dataset a sample from a larger set?}

Yes. The 1,345 clips are curated by applying stratified sampling to 10,024 candidate clips, resulting in a 27:1 reduction. Sampling prioritises action-class coverage and site diversity while accounting for visibility degradation and preserving the natural operational distribution.

\paragraph{Is there information missing?}

Continuous video between sampled frames is not included (only 8 frames per clip). Audio is not included. Worker identity is not recorded and cannot be determined from the imagery. Two of 25 action classes have fewer than 15 instances, limiting per-class statistical claims for those classes.

\paragraph{Are there known errors or noise?}

Known sources of noise include: (1) anchoring bias from VLM pre-fill (+13.9pp on action classification); (2) harmful anchor rate of 3.9\% where annotators accepted incorrect VLM suggestions; (3) spatial context inter-annotator agreement ($\kappa$=0.632). These are quantified through the provenance audit protocol (Section~\ref{sec:audit}).

\paragraph{Does the dataset contain confidential data?}

Faces are blurred with Gaussian filters. Workers appear as 80--200-pixel figures at 7--10 meters of surveillance distance and cannot be individually identified. No names, IDs, or demographic information are recorded.

\paragraph{Could the data be offensive?}

No. The footage shows routine industrial work activities.

\subsection{Collection Process}

\paragraph{How was the data collected?}

Footage was recorded by the plant's existing CCTV security infrastructure (fixed overhead cameras). The research team did not install cameras or direct recording. Clips were extracted using an automated pipeline: FFmpeg for video decoding, YOLO for person detection at 1 fps, and BRISQUE for image quality filtering.

\paragraph{Who was involved in the collection?}

The CCTV system operates continuously as part of plant security. The research team accessed recordings with institutional approval from plant management. No workers were recruited or directed for data collection.

\paragraph{Over what timeframe?}

Source footage spans November 2025 to April 2026 (five months of continuous plant operation).

\paragraph{Were individuals notified?}

Workers operate under existing plant surveillance policies. The CCTV system is part of the standard plant security infrastructure with signage. The research uses only anonymised clips where individuals cannot be identified.

\paragraph{Did subjects consent?}

Workers are not experimental subjects. They perform normal duties under existing security cameras. Individual consent is not applicable because (1) the cameras are pre-existing infrastructure, (2) faces are blurred in the released dataset, (3) individuals cannot be identified from the 7-10 meter overhead viewpoint, and (4) the recent policy of using surveillance recording for safety violation detection is widely circulated.

\paragraph{Was an ethical review conducted?}

Institutional approval was obtained from plant management for the use of the footage for research. No IRB review was required because workers are not experimental subjects and cannot be identified in the released data.

\subsection{Preprocessing, Cleaning, and Labelling}

\paragraph{What preprocessing was done?}

(1) Clip extraction: 15-second segments with $\geq$1 detected person, quality-filtered by BRISQUE score. (2) Frame sampling: 8 frames per clip at fixed temporal positions. (3) Face anonymisation: Gaussian blur applied to all detected faces. (4) Temporal deduplication: near-duplicate or minimal activity or same activity for prolonged duration(working with a tool for 5 minutes) clips are removed.

\paragraph{Was raw data saved?}

Yes. The raw surveillance footage is retained at the facility. The released dataset contains the preprocessed 8-frame representation with face anonymisation applied.

\paragraph{Is the preprocessing software available?}

Yes. The clip extraction pipeline, frame sampling code, and annotation tool are available at \url{https://anonymous.4open.science/r/steelbench-release-0FD6}.

\paragraph{How was labelling done?}

A four-stage pipeline: (1) VLM pre-fill using Qwen3-VL-235B-A22B via API, generating structured annotations; (2) five trained Tier-1 annotators review and correct pre-fills using a custom web interface; (3) two domain experts(junior safety officers from the plant) verify annotations; (4) a senior safety officer adjudicates and conducts a random audit of safety-relevant clips. Each annotator completed a 25-clip calibration session with expert feedback before entering the main queue. Provenance is tracked at each stage, recording which fields were modified and by whom (Section~\ref{sec:audit}).

\subsection{Uses}

\paragraph{Has the dataset been used for any tasks already?}

This paper presents the first use: evaluating 9 VLMs on industrial activity recognition and safety reasoning.

\paragraph{What other tasks could it be used for?}

Worker detection and counting, PPE compliance monitoring, scene understanding under visibility degradation, multi-label classification under natural class imbalance, and studying the effects of model-assisted annotation on benchmark reliability.

\paragraph{Are there tasks it should not be used for?}

The dataset should \emph{not} be used for: (1) autonomous safety monitoring without additional validation, (2) individual worker surveillance or tracking, (3) re-identification of individuals, (4) training production safety systems without domain-specific validation. The CC-BY-NC-4.0 license prohibits commercial use.

\paragraph{How might composition or collection affect future uses?}

The dataset reflects a single integrated steel plant. An integrated steel plant comprises a wide range of industrial units, including chemical plants, power plants, warehouses, workshops, mines, and raw-material-handling facilities, within a large industrial complex. However, cross-facility validation of results across different geographical units has not been conducted. 

\subsection{Distribution}

\paragraph{How is the dataset distributed?}

The dataset is hosted on HuggingFace (\url{https://huggingface.co/datasets/steelbench/SteelBench-release}) with Croissant metadata and Responsible AI (RAI) fields. Evaluation code is available on GitHub (\url{https://anonymous.4open.science/r/steelbench-release-0FD6}).

\paragraph{When will it be released?}

The dataset is available at \url{https://huggingface.co/datasets/steelbench/SteelBench}. A DOI will be assigned through HuggingFace upon acceptance.

\paragraph{What license?}

CC-BY-NC-4.0 (Creative Commons Attribution-NonCommercial 4.0 International). This permits research use with attribution but prohibits commercial deployment.

\paragraph{Are there access restrictions?}

Users must accept the HuggingFace usage terms, which prohibit re-identification attempts and deployment without additional validation. No fees are required.

\subsection{Maintenance}

\paragraph{Who maintains the dataset?}
The dataset creators. 

\paragraph{How can the maintainer be contacted?}

Via the HuggingFace dataset card (\url{https://huggingface.co/datasets/steelbench/SteelBench}) or GitHub repository (\url{https://anonymous.4open.science/r/steelbench-release-0FD6}).

\paragraph{Will the dataset be updated?}

Error corrections and metadata updates will be documented through semantic versioning on HuggingFace. Planned expansion will include more underrepresented classes and additional domains, such as power plants and mines, in future versions. Raw footage has already been collected for these areas, and over the next 5-6 months, we expect to reach 6000-7000 clips with expanded classes and sites. 

\paragraph{How will updates be communicated?}

Through the HuggingFace dataset card changelog and GitHub repository releases.

\paragraph{Can others extend the dataset?}

The annotation tool and evaluation code are released to enable extensions. Community contributions (errata, additional annotations) can be submitted through GitHub issues. Quality assessment of extensions is the responsibility of the contributing parties.

\subsection{Legal and Ethical Considerations}

\paragraph{Were individuals informed about data collection?}

Workers operate under existing plant surveillance policies. The CCTV system is standard plant security infrastructure. Research use involves only anonymised clips where individuals cannot be identified.

\paragraph{Were there ethical review applications?}

Institutional approval was obtained from plant management. No IRB was required because workers are not experimental subjects and cannot be identified.

\paragraph{Could the dataset expose people to harm?}

Risk is very minimal. Faces are Gaussian-blurred, individuals are unidentifiable at surveillance distance (80--200 pixel figures), and no names or IDs are recorded. The CC-BY-NC-4.0 license and its usage terms prohibit re-identification.

\paragraph{Does it unfairly advantage or disadvantage any group?}

The dataset does not record demographic attributes. Annotations describe worker actions and PPE, not individual characteristics. However, models trained on this data inherit any facility-specific biases in work practices or safety enforcement.

\paragraph{Does it comply with GDPR?}

The dataset contains no personally identifiable information. Faces are blurred, and individuals cannot be identified. The data is collected under legitimate institutional interest for workplace safety research.

\paragraph{Does it contain sensitive information?}

No personally identifiable information. The facility layout shown in the frames is already visible in public satellite imagery. Safety violation annotations reference general industrial safety codes, not individual disciplinary records.


\newpage
\section*{NeurIPS Paper Checklist}

\begin{enumerate}

\item {\bf Claims}
    \item[] Question: Do the main claims made in the abstract and introduction accurately reflect the paper's contributions and scope?
    \item[] Answer: \answerYes{}
    \item[] Justification: The abstract states specific findings (capability fragmentation, 42.0pp human--model gap, safety mode clustering, DRS 0.40 ceiling), all supported by experimental results in Section~\ref{sec:results} and ablations in Section~\ref{sec:ablations}. Contributions are enumerated in Section~\ref{sec:introduction}.

\item {\bf Limitations}
    \item[] Question: Does the paper discuss the limitations of the work performed by the authors?
    \item[] Answer: \answerYes{}
    \item[] Justification: Section~\ref{sec:conclusion} discusses limitations. SteelBench is designed as a diagnostic evaluation benchmark, not a training corpus or leaderboard dataset. It tests whether VLMs possess the specific capabilities required for industrial deployment rather than ranking models based on a single accuracy metric. Known limitations include a single-facility scope(all clips from one integrated steel plant), and cross-facility generalisation is untested. Anchoring bias from VLM pre-fill (+13.9pp on action classification), spatial context IAA at ($\kappa$=0.632) and 2 of 25 action classes with $<$15 instances, and 8-frame sampling rather than continuous video. The datasheet (Appendix~\ref{app:datasheet_full}) includes a comprehensive known limitations section.      

\item {\bf Theory assumptions and proofs}
    \item[] Answer: \answerNA{}
    \item[] Justification: This is an empirical evaluation benchmark paper with no theoretical results.

\item {\bf Experimental result reproducibility}
    \item[] Answer: \answerYes{}
    \item[] Justification: Section~\ref{sec:dataset} describes dataset construction. Section~\ref{sec:framework} defines all evaluation metrics with formulas (full definitions in Appendix~\ref{app:metrics}). Section~\ref{sec:results} specifies the 9 evaluated models, prompt design, and evaluation protocol. Evaluation code and all model inference outputs are available at \url{https://anonymous.4open.science/r/steelbench-release-0FD6}.

\item {\bf Open access to data and code}
    \item[] Answer: \answerYes{}
    \item[] Justification: The dataset (1,345 clips with anonymised frames, annotations, and 9-model inference results) is hosted on HuggingFace (\url{https://huggingface.co/datasets/steelbench/SteelBench}) with Croissant metadata and RAI fields. Evaluation code is available on GitHub (\url{https://anonymous.4open.science/r/steelbench-release-0FD6}) under an open-source license. Dataset released under CC-BY-NC-4.0.

\item {\bf Experimental setting/details}
    \item[] Answer: \answerYes{}
    \item[] Justification: Section~\ref{sec:framework} specifies the 9 evaluated models across 4 architectural families, input format (8 frames at 1080p resolution), prompt content (action taxonomy, safety rules, site context), and output schema. Section~\ref{sec:audit} describes the three-tier annotation pipeline and ground truth construction. Appendix~\ref{app:full_results} provides per-model, per-class, per-site, and per-condition breakdowns.

\item {\bf Experiment statistical significance}
    \item[] Answer: \answerYes{}
    \item[] Justification: We report sample sizes for all metrics (289 DA pairs for IAA, 177 proper-chain clips for direction analysis, 102 blind pairs, 150 clips for ablations). Per-class results note where $n<15$ limits interpretation. DRS sensitivity analysis (Appendix~\ref{app:drs_sensitivity}) tests robustness to $\pm$10\% threshold variation.

\item {\bf Experiments compute resources}
    \item[] Answer: \answerYes{}
    \item[] Justification: Annotation VLM inference: Qwen3-VL-235B via DeepInfra API. Evaluation VLM inference: 9 models via commercial APIs (DeepInfra, OpenAI, Anthropic, Google). Clip extraction: NVIDIA A6000 48GB GPU server. No dedicated GPU required for annotation or evaluation. The web interface for annotators, experts, and safety officers is hosted on an AWS EC2 instance, and all VLM inference uses commercial APIs.

\item {\bf Code of ethics}
    \item[] Answer: \answerYes{}
    \item[] Justification: Footage collected under institutional approval from plant management. No individual identification possible at 7-10m surveillance distance (workers appear as 80--200 pixel figures). Face anonymisation (Gaussian blur) applied as an additional safeguard. Workers are not experimental subjects; they perform their normal duties, which are captured by existing security cameras. Section~\ref{sec:conclusion} and Appendix~\ref{app:datasheet_full} discuss ethical considerations.

\item {\bf Broader impacts}
    \item[] Answer: \answerYes{}
    \item[] Justification: Section~\ref{sec:conclusion} discusses positive impact (revealing VLM limitations prevents premature deployment of inadequately evaluated safety monitoring) and risks (surveillance concerns, potential for over-reliance on automated systems). The dataset is explicitly not intended for autonomous deployment, real-time monitoring, or individual worker surveillance.

\item {\bf Safeguards}
    \item[] Answer: \answerYes{}
    \item[] Justification: Face anonymisation applied to all released frames. Dataset licensed CC-BY-NC-4.0 for research only. Access requires acceptance of usage terms prohibiting re-identification attempts and deployment without additional validation. Only 8 sampled frames per clip were released (not a continuous video). Safety rules sourced from the plant's official documentation.

\item {\bf Licenses for existing assets}
    \item[] Answer: \answerYes{}
    \item[] Justification: YOLOv8n (AGPL-3.0) used for person detection during clip extraction. Qwen3-VL-235B accessed via DeepInfra API for pre-annotation. Evaluated models accessed via commercial APIs (OpenAI, Anthropic, Google, DeepInfra). All referenced benchmarks and methods are properly cited.

\item {\bf New assets}
    \item[] Answer: \answerYes{}
    \item[] Justification: SteelBench dataset documented with Gebru datasheet (Appendix~\ref{app:datasheet_full}), Croissant metadata with RAI fields, and dataset card. Released under CC-BY-NC-4.0 license. Evaluation code released separately on GitHub. Appendix~\ref{app:metrics} provides complete metric definitions.

\item {\bf Crowdsourcing and research with human subjects}
    \item[] Answer: \answerYes{}
    \item[] Justification: Annotations produced by 5 trained Tier-1 annotators, 2 domain experts(junior safety officers from plant), and 1 senior safety officer. Annotator identities are anonymised (referred to as annotator\_1 through annotator\_10). Per-annotator statistics reported in Appendix~\ref{app:audit_details}. Annotation quality metrics (IAA, override rate, direction analysis) are reported in Section~\ref{sec:audit}.

\item {\bf Institutional review board (IRB) approvals or equivalent for research with human subjects}
    \item[] Answer: \answerNA{}
    \item[] Justification: Annotators are research team members performing annotation tasks, not experimental subjects. Experts and the Safety Officer are nominated by the plant to support the research. Surveillance footage collected under existing plant security infrastructure with institutional approval from plant management. Workers captured on footage are not experimental subjects, and they perform their normal duties under existing security cameras.

\item {\bf Declaration of LLM usage}
    \item[] Answer: \answerYes{}
    \item[] Justification: Qwen3-VL-235B-A22B-Instruct is used as the VLM pre-annotation tool in the annotation pipeline (Section~\ref{sec:audit}). Nine VLMs are evaluated as the subject of the benchmark (Section~\ref{sec:results}). GPT-4o-mini is used to map free-form descriptions to the action taxonomy in the prompt sensitivity ablation (Section~\ref{sec:ablations}). All LLM usage is a core component of the research methodology, not incidental assistance.

\end{enumerate}

\end{document}